\pdfoutput=1

\documentclass[review]{elsarticle}

\makeatletter
\def\ps@pprintTitle{%
  \let\@oddhead\@empty
  \let\@evenhead\@empty
  \def\@oddfoot{\reset@font\hfil\thepage\hfil}
  \let\@evenfoot\@oddfoot
}
\makeatother

\usepackage{lineno,hyperref}
\modulolinenumbers[5]

\journal{Knowledge-Based Systems}

\usepackage{times}
\usepackage{latexsym}

\usepackage[T1]{fontenc}
\usepackage[utf8]{inputenc}
\usepackage{microtype}
\usepackage{latexsym}
\usepackage{graphicx}
\usepackage{amsmath}
\usepackage{amssymb}
\usepackage{soul,xcolor}
\usepackage{subfig}
\usepackage{float}
\usepackage{multirow}
\usepackage{times}
\usepackage{latexsym}
\usepackage{microtype}
\usepackage{array}
\newcolumntype{H}{>{\setbox0=\hbox\bgroup}c<{\egroup}@{}}
\usepackage{colortbl}
\usepackage{soul,framed}
\usepackage{enumitem}

\usepackage{tikz}
\usetikzlibrary{shapes,snakes,positioning,backgrounds,calc,tikzmark, fit}
\usepackage{pgfplots}
\usepgfplotslibrary{groupplots}

\usepackage{tablefootnote}
\usepackage{natbib}

\newcommand{\multitask}{{\tt EFR-ERC}\textsubscript{multi}}

\newcommand{\cascadeErcEfr}{{\tt ERC$\rightarrow$EFR$_{cas}$}}
\newcommand{\cascadeErcTrueEfr}{{\tt ERC$^{True}\rightarrow$EFR}}

\usepackage{arydshln}

\begin{document}

\begin{frontmatter}

\title{Discovering Emotion and Reasoning its Flip in Multi-Party Conversations using Masked Memory Network and Transformer}

\author {Shivani Kumar}
\author {Anubhav Shrimal}
\author {Md Shad Akhtar}
\author {Tanmoy Chakraborty}

\address {Dept. of CSE, IIIT-Delhi, India}

\begin{abstract}
Efficient discovery of a speaker's emotional states in a multi-party conversation is significant to design human-like conversational agents. During a conversation, the cognitive state of a speaker often alters due to certain past utterances, which may lead to a flip in their emotional state. Therefore, discovering the reasons (triggers) behind the speaker's emotion-flip during a conversation is essential to explain the emotion labels of individual utterances. In this paper, along with addressing the task of emotion recognition in conversations (ERC), we introduce a novel task -- {\bf Emotion-Flip Reasoning} (EFR), that aims to identify past utterances which have triggered one's emotional state to flip at a certain time. We propose a masked memory network to address the former and a Transformer-based network for the latter task. To this end, we consider MELD, a benchmark emotion recognition dataset in multi-party conversations for the task of ERC, and augment it with new ground-truth labels for EFR.  An extensive comparison with five state-of-the-art models suggests improved performances of our models for both the tasks. We further present anecdotal evidence and both qualitative and quantitative error analyses to support the superiority of our models compared to the baselines.
\end{abstract}

\begin{keyword}
Emotion Recognition, Emotion-Flip Reasoning, Multi-Party Conversations
\end{keyword}

\end{frontmatter}

\section{Introduction}
\label{sec:intro}
Early studies in the area of emotion analysis \cite{Ekman92anargument,Picard:1997:AC:265013} primarily focused on emotion recognition in standalone text \cite{abdul-mageed-ungar-2017-emonet,chatterjee-etal-2019-semeval,akhtar:all:in:one:affective} which has been shown to be effective in a wide range of applications such as e-commerce \cite{gupta-etal-2010-emotion}, social media \cite{dini-bittar-2016-emotion,akhtar:cim:emotion}, and health-care \cite{khanpour-caragea-2018-fine}. Recently, the problem of emotion analysis has been extended to the conversation domain -- usually dubbed as {\bf Emotion Recognition in Conversation}, \textit{aka} {\bf ERC} \cite{hazarika-etal-2018-conversational}, where the inputs are no longer standalone; instead, they appear as a sequence of utterances uttered by more than one speaker. The aim of ERC is to  extract the expressed emotion of every utterance in a conversation (or dialogue).

\noindent \paragraph{\bf Motivation} 
A thorough analysis of the ERC task points to multiple important research avenues beyond traditional emotion recognition. One such avenue is the {\em explainability} of the speaker's emotional dynamics within a conversation. Our emotional state changes very often as human beings, which is universal for any conversation among multiple interlocutors. In general, the flip in emotion usually happens due to two factors -- implicit (or external) and explicit (or internal). The implicit factor corresponds to the external or unknown instigator, and it is extremely difficult to identify the exact reason for an emotion-flip. For example, someone's emotion can change without any verbal signal. On the contrary, the explicit factor always has some instigators, e.g., another speaker's visual or verbal signals that trigger the flip in a person's emotion.

Exploring reasons behind emotion-flips of a speaker has wide variety of applications. For example, a dialogue agent can utilize this as feedback for the response generation, as and when, it senses an emotion-flip due to one of its generated responses. A positive emotional-flip (e.g., \textit{\textcolor{orange!60!black}{sadness}} $\rightarrow$ \textit{\textcolor{green!30!black}{joy}}) can be treated as a reward, whereas the system can penalize the agent for a negative emotional-flip (e.g., \textit{\textcolor{blue!60!black}{neutral}} $\rightarrow$ \textit{\textcolor{red!60!black}{angry}}).
A few researchers like {Lin et al.} \cite{lin2019moel} explored the domain of empathetic response generation. In their work, they first captured the user emotions to get an emotion distribution which is used to optimize the response. On similar lines, {Shin et al.} \cite{shin2020generating} assigned a higher reward to a generative model if it generated an utterance that improved the user's sentiment. They proposed a new reward function under the reinforcement learning framework to do so. {Ma et al.} \cite{ma2020survey} highlighted more such systems in their survey paper. {Young et al.} \cite{young2020dialogue} used audio context to train a generative model. They first learnt an auxiliary response retrieval task for audio representation learning followed by discovering additional context by fusing word-level modalities. Other than empathetic response generation, another possible application of identifying the triggers for an emotion-flip is in the domain of affect monitoring. An organization or an individual can reason upon the emotion-flip in a conversation and make an informed decision for a downstream task.

\begin{figure}
    \centering
    \subfloat[Emotion-flip is caused by the previous utterance. Out of five emotion-flips, we show only two of them ($u_5\rightarrow u_7$ and $u_6 \rightarrow u_8$) for aesthetic reasons. Other emotion-flips are $u_1\rightarrow u_3$, $u_2\rightarrow u_4$, and $u_4\rightarrow u_6$ with triggers $u_3$, $u_3$, and $u_6$, respectively.\label{fig:trigger:example:normal}]{
    \includegraphics[width=0.9\textwidth]{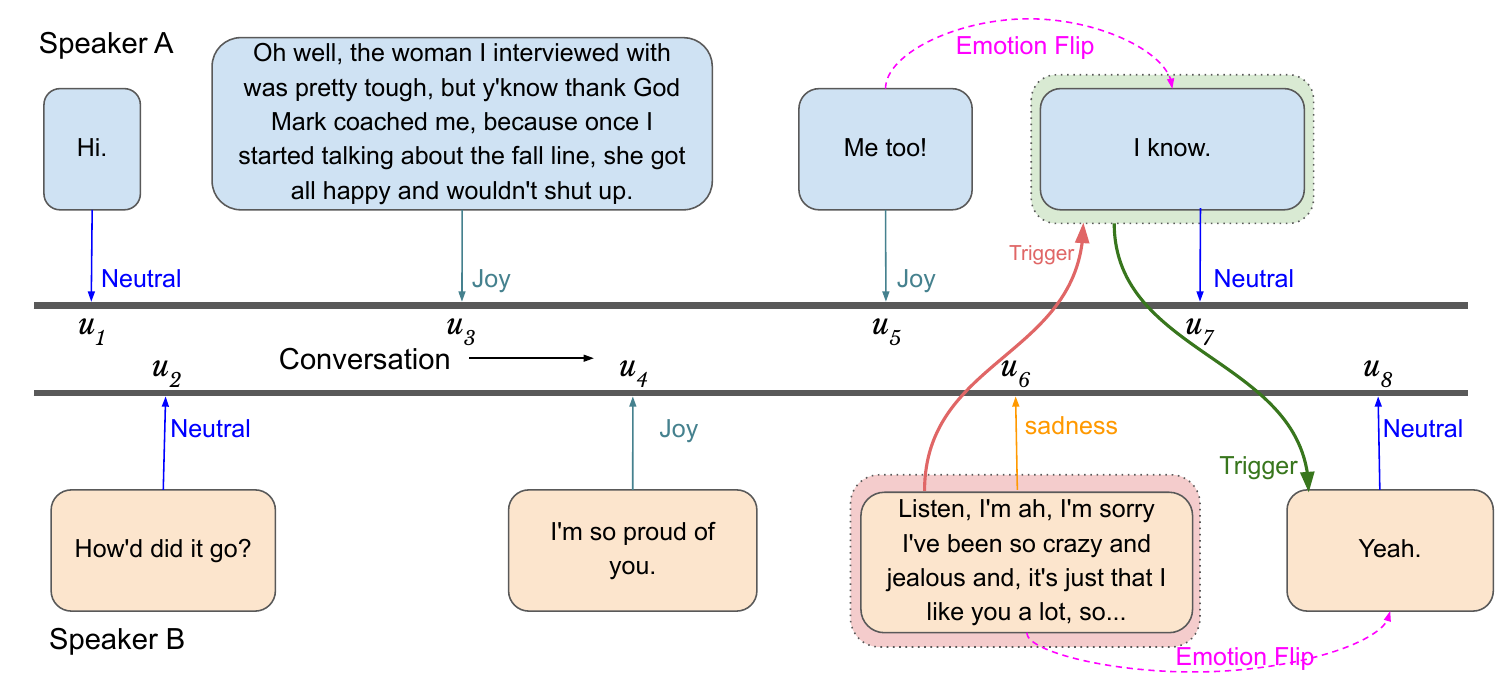}}
    \\
    \subfloat[Emotion-flip is caused by more than one utterance.\label{fig:trigger:example:multiple}]{
    \includegraphics[width=0.9\textwidth]{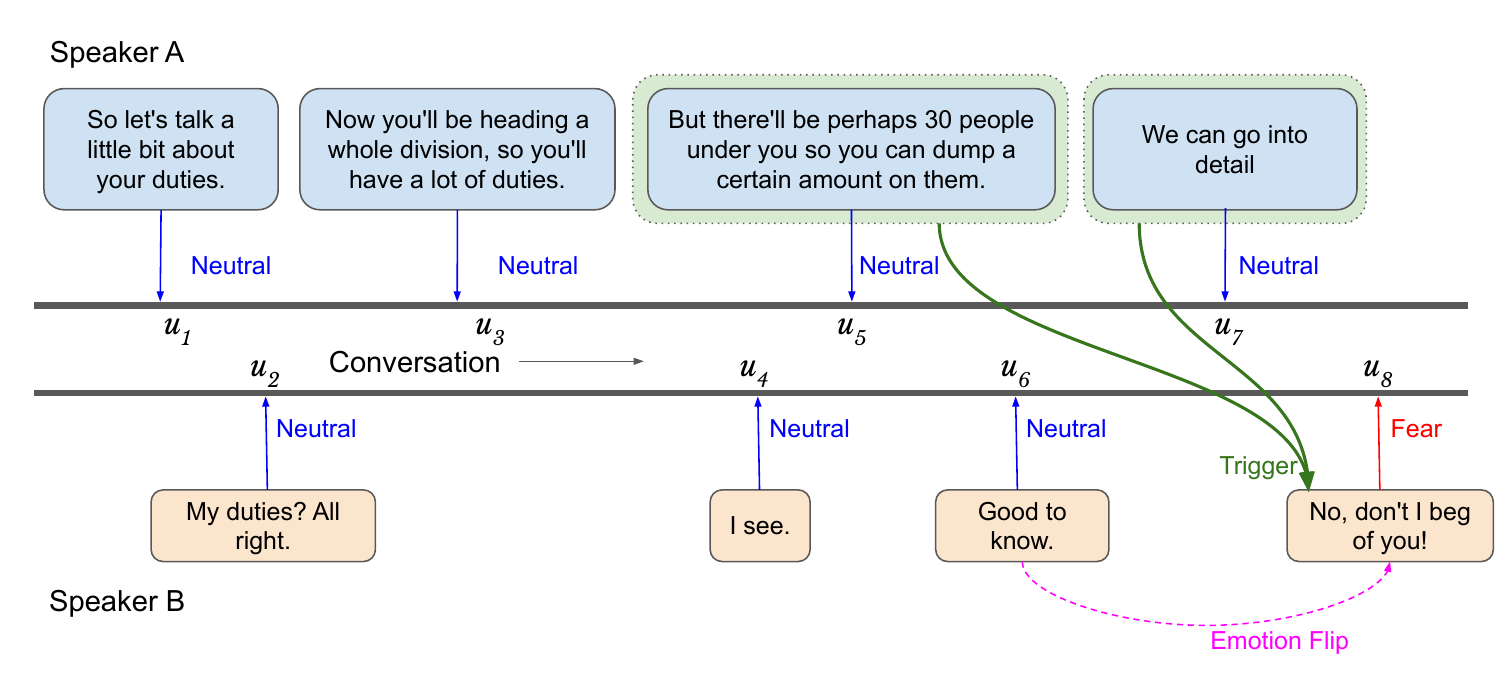}}
    \\
    \subfloat[Emotion-flip in a multi-party conversation.\label{fig:trigger:example:multiparty}]{
    \includegraphics[width=0.58\textwidth]{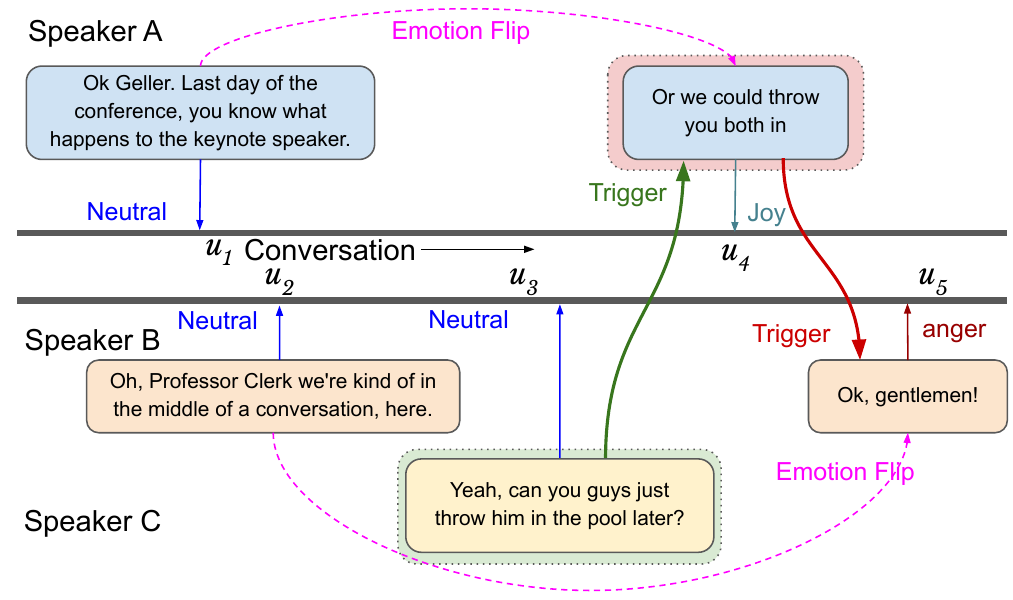}}
    \hspace{1em}
    \subfloat[Self-trigger emotion-flip.\label{fig:trigger:example:self-trigger}]{
    \includegraphics[width=0.35\textwidth]{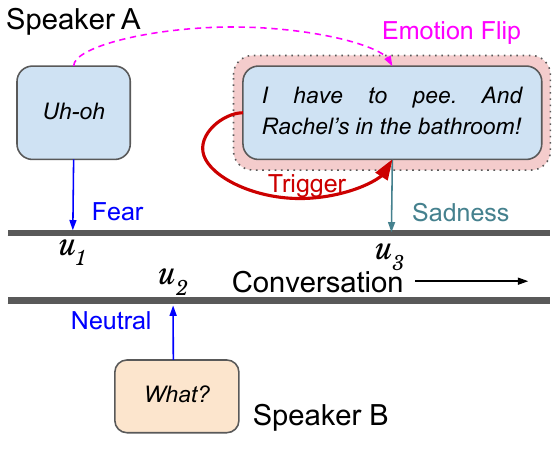}}
    \caption{Examples of emotion-flip reasoning.}
    \label{fig:esr:example}
\end{figure}

\noindent \paragraph{\bf Novel Problem Definition} 
To this end, we introduce a novel problem, called {\bf Emotion-Flip Reasoning}, {\em aka} {\bf EFR}, in conversations. The goal is to find all utterances that trigger a flip in emotion of a speaker within a dialogue. A few example scenarios are presented in Figure \ref{fig:esr:example}. The first dialogue in Figure \ref{fig:trigger:example:normal} exhibits five emotion-flips, i.e., $u_1$ (\textit{\textcolor{blue!60!black}{neutral}}) $\rightarrow u_3$ (\textit{\textcolor{green!30!black}{joy}}), $u_2$ (\textit{\textcolor{blue!60!black}{neutral}}) $\rightarrow u_4$ (\textit{\textcolor{green!30!black}{joy}}), $u_4$ (\textit{\textcolor{green!30!black}{joy}}) $\rightarrow u_6$ (\textit{\textcolor{orange!60!black}{sadness}}), $u_5$ (\textit{\textcolor{green!30!black}{joy}}) $\rightarrow u_7$ (\textit{\textcolor{orange!60!black}{sadness}}), and $u_6$ (\textit{\textcolor{orange!60!black}{sadness}}) $\rightarrow u_8$ (\textit{\textcolor{blue!60!black}{neutral}}); and the utterances which trigger the emotion flips are $u_3$, $u_3$, $u_6$, $u_6$ and $u_7$, respectively.  
Note that some emotion-flips might not be triggered by other speakers in the dialogue; instead, the target utterance can act as a {\em self-trigger}. We show such a scenario in Figure \ref{fig:trigger:example:self-trigger} in which utterance $u_3$ is the only reason for the emotion-flip observed. On the other hand, Figure \ref{fig:trigger:example:multiple} shows a case where more than one trigger is instigating an emotion-flip while Figure \ref{fig:trigger:example:multiparty} presents an example where more than two speakers are involved in the conversation. In such a case, the trigger can come from any of the speakers' utterances.

Formally, the EFR task can be defined as follows: Let $U= [u_1, u_2, ..., u_n]$ be a sequence of $n$ utterances in a dialogue conversation $D$ uttered by $m$ distinct speakers $S = \{s_1, s_2, ..., s_m\}$. As the conversation progresses within a dialogue, these speakers utter their views or feelings in response to the previous utterances. For each utterance $u^{s_j}_i$ in the dialogue, we have an associated emotion $E^{s_j}_i \in$ \{\textit{anger}, \textit{fear}, \textit{disgust}, \textit{sad}, \textit{happy}, \textit{surprise}, \textit{neutral}\}, i.e., $E^{s_j}_i = f_E(u^{s_j}_i)$. Consequently, if the emotion expressed in utterance $u^{s_j}_i$ flips w.r.t the speaker $s_j$'s last utterance $u^{s_j}_l$, there might be a set of associated trigger-utterances $u_k$, $1 \le k \le i$, responsible for the emotion-flip of $s_j$, i.e., $[..., u_k, ...] = f_T(u^{s_j}_i)$. 
In case of no emotion-flip, we associate a `\textit{non-trigger}' label to the current utterance.

\noindent \paragraph{\bf Novel Frameworks for ERC and EFR} We propose a masked memory network based framework for ERC and a Transformer-based model for EFR. Our ERC architecture effectively fuses the dialogue-level global conversational and speaker-level local conversational contexts to learn an enhanced representation for each utterance. Furthermore, we employ a memory network \cite{end2end:mem:net:subakhtar:nips:2015} to leverage the historical relationship among all the previous utterances and their associated emotions as additional information. We hypothesize that the memory content at state $t$ can model the dialogue-level emotional dynamics among the speakers so far. Thus, it can be a vital piece of information for the future utterances corresponding to the states $t+y$, where $y = 1,\cdots,n-t$. On the other hand, we propose a Transformer-based model for EFR. We model the EFR task as an instance-based trigger classification, i.e., for each target (position of emotion-flip) utterance, we predict \textit{trigger/non-trigger} for all the past utterances.

For evaluation, we augment MELD \cite{poria-etal-2019-meld}, a benchmark ERC dataset with ground-truth EFR labels. The resultant dataset, called {\bf MELD-FR}, contains $8,387$ trigger utterances for $5,430$ emotion-flips. We evaluate our models on MELD and MELD-FR for the ERC and EFR tasks, respectively. We also perform extensive ablation and comparative studies with five baselines and different variations of our models and obtain state-of-the-art results for both tasks. A side-by-side diagnostics and anecdotes further explore the errors incurred by the competing models and explain why our models outperform the baselines.

\noindent \paragraph{\bf Contributions} 
Our major contributions are four-fold: 
\begin{enumerate}
    \item[\bf a.] We propose a novel task, called emotion-flip reasoning (EFR), in the conversational dialogue.
    \item[\bf b.] We develop a new ground-truth dataset for EFR, called MELD-FR.
    \item[\bf c.] We benchmark MELD-FR through a Transformer-based model and present a strong baseline for the EFR task.
    \item[\bf d.] We develop a masked memory network based architecture for ERC, which outperforms several recent baselines.
\end{enumerate}
\noindent \paragraph{\bf Reproducibility} The code and dataset are available at \url{https://github.com/LCS2-IIITD/Emotion-Flip-Reasoning}.

\noindent \paragraph{\bf Organization} The rest of the paper is organized as follows. We present the related works in Section \ref{sec:related}. Dataset construction along with the annotation guidelines is presented in Section \ref{sec:dataset}. We describe our proposed methodology in Section \ref{sec:method}. Experimental results and error analyses are discussed in Section \ref{sec:exp}. Finally, we conclude in Section \ref{sec:conclusion} with immediate future directions.

\section{Related Work}
\label{sec:related}
Interpretability of emotion recognition in the linguistic domain is a relatively new research direction, and only a handful of studies have addressed this issue. Lee et al. \cite{lee-etal-2010-text} studied the cause of the expressed emotion in a text, usually coined as `emotion-cause analysis.' The task is to identify a span in the text that causes a specific emotion. For example, in the sentence \textit{`waiting at the airport is awful, but at the end, I get to see my son.'}, the \textit{joy} emotion is caused by the phrase `\textit{I get to see my son}.' In contrast, the \textit{disgust} emotion is caused by the phrase `\textit{waiting at the airport}.' Gui et al. \cite{gui-etal-2016-event} explored the linguistic phenomenon of the emotion cause and developed an SVM-based model for cause extraction. Poria et al. \cite{poria2016deeper} proposed a hybrid model to explain sarcastic sentences.

\begin{figure}
    \centering
    \subfloat[Emotion-cause extraction]{
    \includegraphics[width=0.5\textwidth]{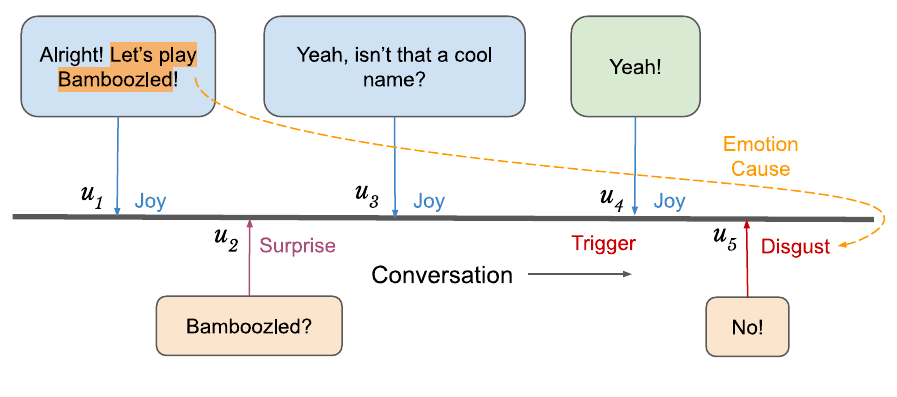}}
    \subfloat[Emotion-flip reasoning]{
    \includegraphics[width=0.5\textwidth]{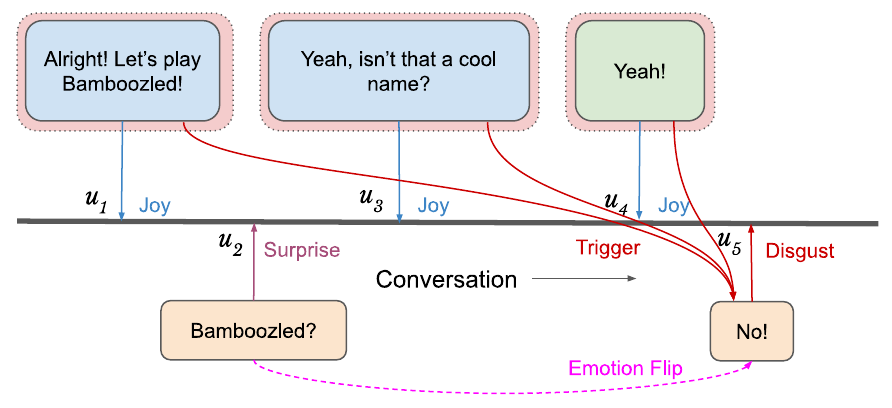}}
    \caption{A sample dialogue to illustrate the difference between emotion-cause extraction in conversation from emotion-flip reasoning.}
    \label{fig:eg_ece_efr}
\end{figure}

At the abstract level, the two tasks, namely emotion-cause analysis \cite{lee-etal-2010-text}, and emotion-flip reasoning may seem related; however, at the surface level, they differ significantly. The emotion-cause analysis aims to find a phrase in the text that provides the clues (or causes) for the expressed emotion. In comparison, our proposed EFR task deals with multiple speakers in a conversational dialogue and aims at extracting the cause (or reason) of an emotion-flip for a speaker. In our case, the triggers are one or more utterances from the dialogue history as opposed to phrases. Figure \ref{fig:eg_ece_efr} illustrates an example from MELD-FR with annotated EFR and emotion-cause labels. It can be observed that the reason behind the emotion \textit{disgust} in utterance $u_5$ is the name of the game which was mentioned in utterance $u_1$. On the other hand, the emotion-flip from \textit{surprise} to \textit{disgust} (from utterances $u_2$ to $u_5$) was triggered by the utterances $u_1$, $u_3$, and $u_4$. 

Emotion knowledge is a piece of vital information in understanding the actual meaning behind a statement. A lot of studies have been done in this field. {Mencattini et al.} \cite{MENCATTINI201468} tried to solve the problem of emotion detection in speech by using PLS regression model, optimized according to specific features selection procedures. Researchers also used other modalities like visual cues and electroencephalography (EEG) signals to obtain the emotional knowledge of the actor. {Zhang et al.} \cite{ZHANG2016248} suggested the use of a variant of the evolutionary firefly algorithm for feature optimization followed by various classical supervised classifiers. {Cui et al.} \cite{CUI2020106243} used EEG signals and learned an end-to-end Regional-Asymmetric Convolutional Neural Network (RACNN) for emotion recognition, which consists of temporal, regional, and asymmetric feature extractors. However, all these models mentioned here work for standalone instances that do not have any context. In the current work, we tackle the task of emotion recognition in conversations, which heavily depends on contextual information.

Emotion recognition in conversation is an active field of study, and several interesting models have been proposed for extracting emotion of every utterance 
\cite{hazarika-etal-2018-conversational,hazarika-etal-2018-icon, hazarika2019emotion,zhong-etal-2019-knowledge,ghosal-etal-2019-dialoguegcn,li2020bieru}.
Hazarika et al. \cite{hazarika-etal-2018-conversational} proposed a Conversational Memory Network (CMN) to leverage the contextual utterances in a dyadic conversation for emotion recognition. The authors showed that maintaining the conversational history in a memory helped CMN in predicting emotions more precisely. In comparison, ICON \cite{hazarika-etal-2018-icon} models the self and inter-speaker influences in the conversation for emotion recognition. Similar to the CMN architecture, ICON also incorporates a memory network to exploit the contextual information within a dialogue.
Ghosal et al. \cite{ghosal-etal-2019-dialoguegcn} aimed to address the issues of context propagation for emotion recognition through an application of graph convolutional network. The inter-speaker dynamics are effectively handled through a complete graph, where each vertex represents an utterance in the dialogue, and the edge-weight reflects the correspondence between two vertices (or utterances). Zhong et al. \cite{zhong-etal-2019-knowledge} proposed to infuse the commonsense knowledge to improve the prediction capability of the ERC model. They argued that the model leverages the knowledge to enrich the hidden representation. Zhang et al. \cite{zhang2019quantum} proposed to capture the intra-utterance and inter-utterance interaction dynamics. They used quantum-inspired interactive networks, which leverage the mathematical formalism of quantum theory and the LSTM network, to learn such interaction dynamics. Wang et al. \cite{wang2020contextualized} modeled the ERC task as sequence tagging. Their architecture used transformer, LSTMs, and CRF to perform the ERC task. {Li et al.} \cite{li2020bieru} used a generalized neural tensor block followed by a two-channel
classifier for performing context compositionality and sentiment classification, respectively.\\

\noindent {\bf How are our methods different?} Following the effective utilization of the memory network for emotion recognition, we also adapt it to leverage the emotional relationship among several interlocutors in this work. However, in contrast to CMN and ICON, we employ it to supplement the global and local emotional dynamics in dialogues captured through a series of recurrent layers. We also handle multiple speakers instead of dyadic conversations, as addressed in many of the existing studies. Moreover, our study advances on the path of explainability by mining the reason behind an emotion-flip of a speaker. {\em To the best of our knowledge, such study has not been carried out in the context of conversation}.

\section{Dataset Construction}\label{sec:dataset}
We employ a recently released dataset, called MELD \cite{poria-etal-2019-meld} and extend it for our EFR task. It consists of $13,708$ utterances spoken by multiple speakers across $1,433$ dialogues from the popular sitcom \textit{F.R.I.E.N.D.S.}\footnote{\url{https://www.imdb.com/title/tt0108778/}} Each utterance has an associated emotion label representing one of Ekman's six basic emotions: \textit{anger}, \textit{fear}, \textit{disgust}, \textit{sadness}, \textit{joy}, and \textit{surprise} along with a label for no emotion, i.e., \textit{neutral}. Though MELD is a multi-modal dataset, in this work, we employ textual modality only.

For EFR, we augment MELD with new ground-truth labels (dubbed as {\bf MELD-FR}). For this, we take inspiration from the \textit{Cognitive Appraisal Theory} by Richard Lazarus \cite{lazarus1984stress} which states that emotions are a result of our evaluations or appraisals of an event. We extend the concept of appraisals and try to identify these in our dialogue instances. We consider the utterances that contain possible appraisals as \textbf{\textit{triggers}}. We employed three annotators who had vast experiences in the ERC task\footnote{However, the annotation process can easily be carried out by anyone if one follows the annotation guidelines we have provided.}. Below, we explain the steps carried out for the EFR annotation.

\begin{enumerate}
    \item For each speaker $s_j$, we identify their utterances $u^{s_j}_{i}$ in a dialogue where a flip in emotion has occurred, i.e., the speaker's last emotion and the current emotion are different.
    \item For each identified utterance $u^{s_j}_i$, we analyse the dialogue context and mark all the utterances $u_k$ (where $1 \le k \le i$) as triggers that are responsible for the emotion-flip in utterance $u^{s_j}_i$.
    For some of the cases, the reason for the emotion-flip of a speaker was not apparent in the dialogue, and the flip was self-driven by the speaker. We leave such cases and do not mark any triggers for them. 
\end{enumerate}

\noindent \paragraph{\bf Annotation Guidelines} For annotating triggers, we define a set of guidelines as furnished below. We define a \textbf{\textit{trigger}} as any utterance in the contextual history of the target utterance (the utterance for which the trigger is to be identified) that follows the following properties:
\begin{enumerate}
    \item The whole utterance or a part of utterance directly influences a change in emotion of the target speaker.
    
    \item The utterance can be uttered by a different speaker or the target speaker.
    
    \item The target utterance can also be classified as a trigger utterance if it contributes to the emotion-flip of the target speaker. For example, if a person's emotion changes from \textit{neutral} to \textit{sad} because of some sad message that she is conveying herself, then the target utterance is the one responsible for the shift.
    
    \item There can be more than one trigger responsible for a single emotion-flip.
    
    \item Since we deal with textual data only, it is possible that the reason behind an emotion-flip is not evident from the data (for example, when the flip occurs due to a visual stimulus). In such cases, no utterance can be marked as a trigger.
\end{enumerate}

We calculate the alpha-reliability inter-annotator agreement \cite{krippendorff2011computing} between each pair of annotators, $\alpha_{AB}=0.824$, $\alpha_{AC}=0.804$, and $\alpha_{BC}=0.820$. To find out the overall agreement score, we take the average score, $\alpha=0.816$. Figure \ref{fig:esr:example} shows a few example scenarios from our dataset. While Figures \ref{fig:trigger:example:normal}, \ref{fig:trigger:example:multiple}, and \ref{fig:trigger:example:self-trigger} illustrate the case when two participants are involved in a conversation, Figure \ref{fig:trigger:example:multiparty} shows an example where more than two speakers are involved. For our work, we considered only those dialogues where speakers experience at least one emotion-flip. After removing dialogues with no emotion-flip, we were left with $834$ dialogues in the training set, which account for $4,001$ utterances with emotion flips. These dialogues were annotated by three annotators according to the above guidelines for identifying triggers. Among three annotators, two of the annotators were male, and one was female. All of them were researchers with 3-10 years of experience. They belong to the age group of 30-40. Though we employed three expert annotators in our annotation phase, the process does not require experts (linguistics, social scientists, etc.). Since the annotation guidelines that we provide above are very generic, they can easily be extended to other dialogue datasets by crowdsourcing. We wanted to prepare our labeled dataset as accurately as possible as it would be the first dataset of its kind.

\begin{figure}[t]
\centering
\includegraphics[width=0.8\textwidth]{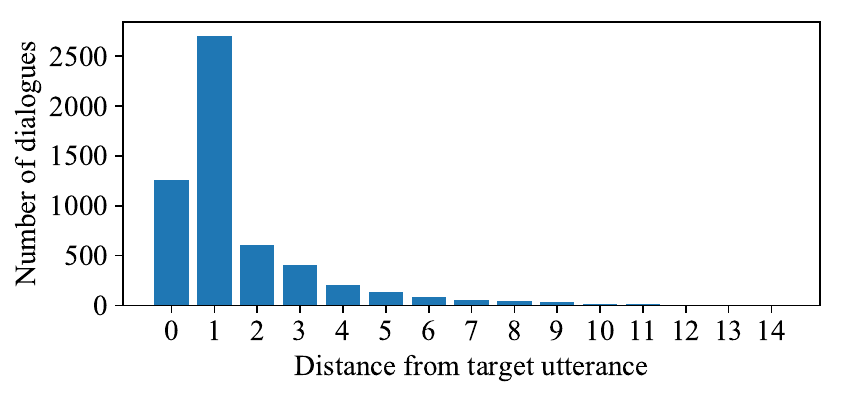}
\captionof{figure}{Distribution of triggers w.r.t their distance from the target utterance.}
\label{fig:eda_trig}
\end{figure}

Similarly, we obtained the trigger annotations for the development and test sets. We show a brief statistic of the datasets in Table \ref{tab:dataset}. The resultant dataset, called MELD-FR, contains $8,387$ trigger utterances for $5,430$ emotion-flips. We also show the EFR trigger distribution considering their distance from the target utterance in Figure \ref{fig:eda_trig}. We observe that for the majority of the emotion-flips, the triggers appear in the past few utterances only. 
\begin{table}[t]
\centering
\centering
    \subfloat[ MELD dataset for Emotion Recognition in Conversation (ERC)]{
    \resizebox{\textwidth}{!}
    {
    \begin{tabular}{c|c|c|c|c|c|c|c|c}
     \multirow{2}{*}{\bf Split} & \multicolumn{7}{c|}{\bf Emotions} & \multirow{2}{*}{\bf Total} \\ \cline{2-8}
     & \bf Disgust & \bf Joy & \bf Surprise & \bf Anger & \bf Fear & \bf Neutral & \bf Sadness & \\ \hline \hline
         \bf Train  & 225 & 1466 & 1021 & 911 & 229 & 3702 & 576 & 8130 \\
         \bf Dev & 20 & 156 & 144 & 126 & 39 & 395 & 97 & 977 \\
         \bf Test & 61 & 325 & 238 & 283 & 42 & 943 & 169 & 2061 \\ \hline
    \end{tabular}}}
    
    \subfloat[ MELD-FR dataset for Emotion-Flip Reasoning (EFR)]{
{
\begin{tabular}{c|c|c|c}
         \bf Split & \bf \#Dialogue with Flip & \bf \#Utterance with Flip & \bf \#Triggers  \\ \hline \hline 
         \bf Train & 834 & 4001 & 6740 \\
         \bf Dev & 95 & 427 & 495 \\
         \bf Test & 232 & 1002 &  1152  \\ \hline
    \end{tabular}}}
    \caption{Statistics of the datasets. We only consider those dialogues from the original MELD dataset where there is at least one emotion-flip. This step removed $271$ dialogues from MELD, resulting $1,161$ dialogues (see Table \ref{tab:erc:results} for the abbreviations of the emotion labels).}
    \label{tab:dataset}
\end{table}

\begin{table}[t]
\centering
\resizebox{\textwidth}{!}{%
\begin{tabular}{c|l|c:c:c:c:c:c:c|c|}
\multicolumn{2}{c}{} & \multicolumn{8}{c}{\textbf{Target}} \\
\cline{3-10}

\multicolumn{1}{c}{} & \multicolumn{1}{c}{} & \multicolumn{1}{|c|}{\textbf{Disgust}} & \multicolumn{1}{c|}{\textbf{Joy}} & \multicolumn{1}{c|}{\textbf{Surprise}} & \multicolumn{1}{c|}{\textbf{Anger}} & \multicolumn{1}{c|}{\textbf{Fear}} & \multicolumn{1}{c|}{\textbf{Neutral}} & \multicolumn{1}{c|}{\textbf{Sadness}} & \multicolumn{1}{c|}{\textbf{Total}} \\ \cline{2-10}

\multirow{9}{*}{\rotatebox{90}{\bf Source}} & \textbf{Disgust} & 0 & 24 & 30 & 47 & 6 & 76 & 13 & 196 \\ \cline{2-2} \cdashline{3-10}
\multicolumn{1}{c|}{} & \textbf{Joy} & 34 & 0 & 169 & 86 & 42 & 665 & 81 & 1077\\ \cline{2-2} \cdashline{3-10}
\multicolumn{1}{c|}{} & \textbf{Surprise} & 39 & 186 & 0 & 137 & 32 & 400 & 70 & 864\\ \cline{2-2} \cdashline{3-10}
\multicolumn{1}{c|}{} & \textbf{Anger} & 37 & 96 & 104 & 0 & 20 & 318 & 99 & 674\\ \cline{2-2} \cdashline{3-10}
\multicolumn{1}{c|}{} & \textbf{Fear} & 7 & 20 & 23 & 45 & 0 & 87 & 27 & 209\\ \cline{2-2} \cdashline{3-10}
\multicolumn{1}{c|}{} & \textbf{Neutral} & 84 & 616 & 487 & 370 & 103 & 0 & 257 & 1917\\ \cline{2-2} \cdashline{3-10}
\multicolumn{1}{c|}{} & \textbf{Sadness} & 17 & 78 & 60 & 72 & 28 & 238 & 0 & 493\\ \cline{2-10}
\multicolumn{1}{c|}{} & \textbf{Total} & 218 & 1020 & 873 & 757 & 231 & 1784 & 547\\ \cline{2-9}
\end{tabular}%
}
\caption{Directionality for emotion flips. The value at cell ($i,j$) denotes the number of emotion-flip from emotion $i$ to emotion $j$.}
\label{tab:ef-direction}
\end{table}

After the annotation process, we analyze the directionality of the emotion-flips in MELD-FR. Table \ref{tab:ef-direction} shows the statistics of the emotion-flip from the source emotion (row) to the target emotion (column). We make some interesting observations. We consider the set of emotion- {\textit{joy} and \textit{surprise}} as `positive' and the set- {\textit{disgust}, \textit{fear}, \textit{anger}, and \textit{sadness}} as `negative' emotions. We analyse the emotion flips based on these positive and negative emotion sets.
There are in total 2612 emotion flips which result in an emotion from the positive set,
whereas 2818 emotion flips result in an emotion from the negative set. Out of these flips, the most prominent emotion-flip pairs are \textit{neutral} 
to \textit{joy} (616) when the resultant (or target) emotion is positive, and \textit{neutral} to \textit{anger} (370) when the target is negative. We also analyse the  flips occurring from a positive to positive emotions (e.g. \textit{joy} to \textit{surprise}) or from negative to negative emotions (e.g. \textit{anger} to \textit{fear}). Within the positive class, the most emotion flips are observed for the pair \textit{surprise} to \textit{joy} (186), while \textit{anger} to \textit{sadness} flip (99) prevails the negative class. Most of the emotion flips that result in a positive emotion originate from \textit{neutral} (1103), while most  emotions flips that result in a negative emotion originate from \textit{joy} (907). When the target emotion of the emotion-flip is a positive emotion, it is mostly \textit{joy} (1020), whereas for the negative case, it is mostly \textit{anger} (757).

We also observe that the most frequent reasons for positive emotion-flip are excitement, cheer, or being impressed by someone else. For negative emotion-flip, awkwardness, loss, or being annoyed are the frequent reasons.

\section{Proposed Methodology}
\label{sec:method}
\begin{figure}[ht!]
    \centering
    \includegraphics[width=\columnwidth]{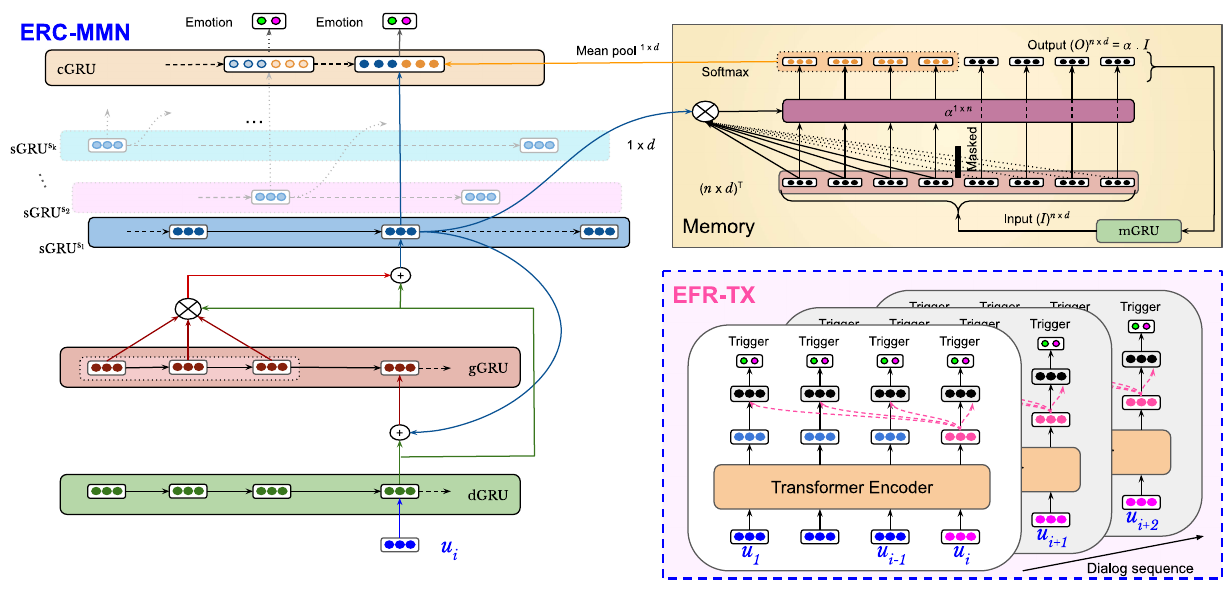}
    \caption{The proposed masked memory network based ({\tt ERC-MMN}) and Transformer-based ({\tt EFR-TX}) models for ERC and EFR, respectively. {\tt ERC-MMN} takes a dialogue (a sequence of  utterances) and aims to predict the emotion of each utterance in order. {\tt EFR-TX} models each instance as a tuple $<$\textcolor{blue}{\em past utterances as trigger candidates}, \textcolor{magenta!100!black!100}{\em target utterance as the location of emotion-flip}$>$.}
    \label{fig:architecture}
\end{figure}

In this section, we describe our proposed models in detail. Since the input-output mapping of the two tasks is different, we model them differently. For ERC, at each time step $t$, we learn an emotion label for  utterance $u_{t}$; whereas, in EFR, we make a sequence of predictions corresponding to each previous utterance $u_i$, where $i \le t$, denoting the trigger/reason behind  emotion-flip at the target utterance $u_t$. We employ an utterance-level memory network for ERC, and an instance-level Transformer-based \cite{vaswani2017attention} encoder for EFR. Figure \ref{fig:architecture} presents a high-level architecture of both the models. The remaining section elaborates on the individual models.

\subsection{\bf Emotion-Flip Reasoning - Transformer (\texttt{EFR-TX})}
We model the emotion-flip reasoning task as an instance-based classification problem. At first, we define an EFR instance as a sequence of utterances, $u_1, u_2, ..., u_t$, where the aim is to identify a set of triggers for the target (last) utterance $u_t$. Intuitively, it can be observed that the triggers for the target utterance $u_t$ must belong to the utterance set $u_1, u_2, ..., u_t$. Thus, we classify each utterance in the instance as trigger/non-trigger for the target utterance $u_t$. We repeat the same process for each utterance in the dialogue; however, if no emotion-flip is observed for the target utterance $u_t$ in the training set, we do not process the error gradients in the backward pass.
We show the instance creation process for all the example dialogues (as shown in Figure \ref{fig:esr:example}) in Table \ref{tab:instance:EFR}.
In the first example (Figure \ref{fig:trigger:example:normal}), there are five utterances, $u_3$, $u_4$, $u_6$, $u_7$ and $u_8$, where the emotion of the speakers has flipped; thus, we have created five instances for each target emotion-flip; the corresponding trigger utterances are $u_3$, $u_3$, $u_6$, $u_6$ and $u_7$, respectively. In the second one (Figure \ref{fig:trigger:example:multiple}), we show an instance where triggers can come from multiple utterances. Here, the emotion of the speaker has flipped once in utterance $u_8$, and the triggers were identified as the utterances $u_5$ and $u_7$. We also show an example when more than two interlocutors are involved in a conversation (Figure \ref{fig:trigger:example:multiparty}). In the particular example, there are two emotion flips at utterances  $u_4$ and $u_5$. The triggers utterances for the same are $u_3$ and $u_4$, respectively. In the last example (Figure \ref{fig:trigger:example:self-trigger}), we show an instance where the emotion-flip is a result of a self trigger, i.e., for the emotion-flip observed at utterance $u_3$, the trigger is the same utterance itself.

\begin{table}[t!]
    \centering
    \begin{tabular}{l|c|l|l}
         & \bf Target $u_t$ & \bf Instance & \bf Trigger labels\\ \hline \hline
         \multirow{5}{*}{Figure \ref{fig:trigger:example:normal}} & \textcolor{red}{$u_3$} & \{$u_1$, $u_2$, \textcolor{red}{$u_3$}\} & \{0, 0, \textcolor{red}{1}\} \\
         & \textcolor{red}{$u_4$} & \{$u_1, u_2, \textcolor{red}{u_3}, u_4$\} & \{0, 0, \textcolor{red}{1}, 0\} \\
         & \textcolor{red}{$u_6$} & \{$u_1, u_2, u_3, u_4, u_5, \textcolor{red}{u_6}$\} & \{0, 0, 0, 0, 0, \textcolor{red}{1}\} \\
         & \textcolor{red}{$u_7$} & \{$u_1, u_2, u_3, u_4, u_5, \textcolor{red}{u_6}, u_7$\} & \{0, 0, 0, 0, 0, \textcolor{red}{1}, 0\} \\
         & \textcolor{red}{$u_8$} & \{$u_1, u_2, u_3, u_4, u_5, u_6, \textcolor{red}{u_7}, u_8$\} & \{0, 0, 0, 0, 0, 0, \textcolor{red}{1}, 0\} \\ \hline
         
         \multirow{1}{*}{Figure \ref{fig:trigger:example:multiple}} & \textcolor{red}{$u_8$} & \{$u_1, u_2, u_3, u_4, \textcolor{red}{u_5}, u_6, \textcolor{red}{u_7}, u_8$\} & \{0, 0, 0, 0, \textcolor{red}{1}, 0, \textcolor{red}{1}, 0\} \\ \hline
        
        \multirow{2}{*}{Figure \ref{fig:trigger:example:multiparty}} & \textcolor{red}{$u_4$} & \{$u_1$, $u_2$, \textcolor{red}{$u_3$}, $u_4$\} & \{0, 0, \textcolor{red}{1}, 0\} \\
         & \textcolor{red}{$u_5$} & \{$u_1, u_2, u_3, \textcolor{red}{u_4}, u_5$\} & \{0, 0, 0, \textcolor{red}{1}, 0\} \\ \hline
        
        \multirow{1}{*}{Figure \ref{fig:trigger:example:self-trigger}}& \textcolor{red}{$u_3$} & \{$u_1$, $u_2$, \textcolor{red}{$u_3$}\} & \{0, 0, \textcolor{red}{1}\} \\ \hline
    \end{tabular}
    
    \caption{Instance creation corresponding to the dialogues in Figure \ref{fig:esr:example} for the EFR task.}
    \label{tab:instance:EFR}
\end{table} 

After compiling the EFR instances, we employ the Transformer model for classification. We obtain the encoder output $h_i$ for each utterance of an instance, and concatenate it with the encoder output of the target utterance $h_t$, i.e., $\forall i, \hat{h}_i = h_i \oplus h_t$. Since emotion-flip reasoning has a strong correspondence with the emotion label, we supplement each contextual utterance with its emotion label to learn an appropriate representation for the trigger classification. Subsequently, we classify each utterance as trigger/non-trigger for the target utterance.

\subsection{\bf Emotion Recognition in Conversation - Masked Memory Network (\texttt{ERC-MMN})}
Recall that a dialogue $D$ can have $n$ utterances spoken by $m$ distinct speakers, and each of these utterances has an associated emotion label $E$. Unlike EFR, we model ERC as the sequence-labeling problem, where for each utterance in the dialogue sequence, we predict its corresponding label.
In our model, we employ $m$ separate speaker-level forward GRUs (sGRU$^{s_j}$)\footnote{To model the natural conversation, we did not account for any future context anywhere in the architecture.} to capture the utterance pattern of each speaker $s_j \in S$. The hidden representations of each sGRU$^{s_j}$ are arranged in the dialogue order and subsequently fed to cGRU for emotion recognition.

For each speaker $s_j \in$ S, we compute a $d$-dimensional speaker-level hidden representation as follows: 
\begin{equation}
    [\bar{h}^{s_j}_1,.., \bar{h}^{s_j}_k,.., \bar{h}^{s_j}_p] = \text{sGRU}^{s_j}(v^{s_j}_1,.., v^{s_j}_k,.., v^{s_j}_p) \nonumber
\end{equation}
where $v^{s_j}_k \in \text{V}^{s_j}$ denotes the utterance spoken by speaker $s_j$ in the dialogue, and $p$ is the number of utterances uttered by  speaker $s_j$. Evidently, 
\begin{equation}U~=~\forall_{s_j \in S} \quad union(\text{V}^{s_j})\nonumber \end{equation}

{Using stacked GRUs, our model becomes agnostic to the number of speakers present in the dialog.} Note that the modeling of a speaker-level GRU, e.g., $s$GRU$^{s_j}$, is in isolation with other speaker-level GRUs (i.e., $s$GRU$^{s_k}$, where $j \ne k$). However, a natural conversation does not happen in isolation; therefore, to provide a mechanism for the interaction among the speakers and to model the natural conversation, we leverage the dialogue-level context in the learning process of the speaker dynamics. The dialogue-level context is maintained through a global GRU (gGRU) which is shared across all the speakers within a dialogue. For each utterance $u^{s_j}_i$, we compute an association with the previous global state gGRU$_{[u_1, ..., u_{i-1}]}$ through an attention mechanism. It ensures that the current utterance is aware of the dialogue-level context. The context-aware attended representation is then forwarded to sGRU$^{s_j}$ to learn the speaker-specific conversation dynamics. Mathematically,
\begin{equation}
v^{s_j}_k = \text{Attention}(\text{gGRU}_{[u_1, ..., u_{i-1}]}, u^{s_j}_i) \oplus u^{s_j}_i \nonumber
\end{equation}
where $u^{s_j}_i$ denotes the $i^{th}$ utterance in the dialogue sequence, and $v^{s_j}_k$ denotes the corresponding $k^{th}$ utterance in the speaker sequence $s_j$. In parallel, the attended vector is consumed by gGRU to update the global state of the dialogue, i.e.,  
$$\text{gGRU}_{[u_1, ..., u_i]} = \text{gGRU}([u_1, ..., u_{i-1}], v^{s_j}_k, d^{s_j}_i)$$
where $d^{s_j}_i$ represents the hidden representation of  utterance $u_i$ in the dialogue sequence.

\paragraph{\bf Masked Memory Network}
To learn the dialogue conversation efficiently, the role of each utterance $u_i$ in the dialogue needs to be carefully examined. Some utterances have lesser importance in the dialogue context, whereas others have a long-lasting effect. In general, there is a higher chance that important utterances may participate in predicting emotions for multiple utterances. In our manual analysis of the dataset, we also observe a reasonable correspondence between a few utterances and the emotional labels for multiple utterances in the dialogue. Hence, we hypothesize that the information regarding these few utterances may be exploited by the future utterances $u_l$, $i < l$, in the dialogue for emotion prediction. To simulate this, we employ a memory network \cite{end2end:mem:net:subakhtar:nips:2015} that maintains the information captured during the previous states. At each state $t$,\footnote{The memory state $t$ corresponds to the utterance $u_i$, $t+1$ to $u_{i+1}$, and so on.} the memory network learns over the state $t-1$ memory content through a forward-GRU (mGRU) and updates it according to the current query $q_t$. The updated memory at state $t$ is then utilized by the network in the emotion recognition for the utterance $u_i$ (represented by the query $q_t$). Furthermore, it also acts as input for the next state $t+1$. We argue that the memory content accumulated over the emotion labels reveals the relationship among previous utterances, and the future utterances leverage it to establish their relationships with the previous utterances.

Here, we employ a masked interactive attention mechanism \cite{IAN:2017} to incorporate the information regarding the current query. For each state $t$, we compute the masked attention weights $\beta_t$ considering input $I_t$ ($I_t= \text{mGRU}(O_{t-1})$) and the query $q_t = \bar{h}_i$. Subsequently, the attended vector is computed for each hidden representation of input $I_t$. Since the masked attention weight $\beta_t$ signifies the probability over the first $t$ input hidden representations (i.e., $I_1, I_2, ..., I_t$) and $\sum \beta_t = 1$, we compute the attended vector for the first $t$ hidden representations only and bypass the rest of the input representations (i.e., $I_{t+1}, ..., I_n$). The two sets of representations, i.e., $t$ attended and $n-t$ bypassed, form the memory output $O_t$.

The attended vectors, i.e., $O_t[1..t]$, are then utilized as the memory context at state $t$ for the subsequent processing corresponding to the utterance $u_i$. We apply mean-pooling over $O_t[1..t]$ to compute $\bar{o}_i$, and concatenate it with the enriched speaker-dialogue hidden representation $\bar{h}_i$ for the final predictions, i.e.,
$E_i = Softmax(\bar{o}_i \oplus \bar{h}_i)$.

\section{Model Evaluation}
\label{sec:exp}
We evaluate our models on MELD-FR. Figure \ref{fig:eda_trig} shows the distribution of triggers based on their distance from the target utterance. We notice that most of the triggers are the utterances that are spoken just before the target. This phenomenon corresponds to the natural conversations where an emotion-flip occurs immediately after a trigger statement is said. However, there are cases where the trigger lies beyond the last utterance of the target speaker. After analyzing the distribution carefully, we restrict the {\tt context\_size = 5} for the experiments involving instance-level EFR classification.
\subsection{Baseline Methods}
In this work, we employ a set of baseline methods for the comparative study \cite{hazarika-etal-2018-conversational,hazarika-etal-2018-icon, ghosal-etal-2019-dialoguegcn,jiao2019real,vinyals2015pointer}
\begin{itemize}
    \item \textbf{CMN} \cite{hazarika-etal-2018-conversational}\textbf{:} It utilizes memory networks to store the speaker-level contextual history within a dialogue. The authors showed that maintaining the conversational history in a memory helped CMN in predicting emotions more precisely. They also used these memories in capturing inter-speaker dependencies.
    
    \item \textbf{ICON} \cite{hazarika-etal-2018-icon}\textbf{:} It maintains a memory network to preserve the interaction between the \textit{self} and \textit{inter-speaker} influences in dyadic conversations. It models this interaction into the global memory in a hierarchical way. Finally, the memory is used as a contextual summary which aid in predicting the emotional labels.
    
    \item \textbf{DGCN} \cite{ghosal-etal-2019-dialoguegcn}\textbf{:} It models the inter-speaker dynamics in a dialogue via a graph convolutional network. This work also leverages the self and inter-speaker dependencies of the participants for modeling conversations. By using graphs, the authors claim to have modeled context propagation in an efficient way.
    
    \item \textbf{AGHMN} \cite{jiao2019real}\textbf{:} It incorporates an attention GRU mechanism that controls the flow of information through a modified GRU cell based on the attention-weights, computed over the historical utterances in a dialogue.
    
    \item \textbf{Pointer Network} \cite{vinyals2015pointer} \textbf{:} They are often used to generate output sequence when the length of output sequence depends on the length of the input sequence. Pointer networks have been applied to solve various combinatorial optimization and search problems such as Convex hull, and travelling salesman problem. Here, we use it to map our input sequence of utterances of a dialogue into a sequence of emotions or triggers.
\end{itemize}

These baselines are readily suitable for ERC, and a few of them reported their performance on the MELD dataset. However, \textit{we reproduced the results of these baselines on MELD-FR using the official repository of each baseline.}\footnote{For consistency, we evaluate our model on MELF-FR instead of original MELD for both ERC and EFR tasks. MELD-FR differs from MELD since it does not contain dialogues with no emotion-flip.} On the other hand, by definition, for each utterance $u_i$, EFR aims to predict a classification (trigger/non-trigger) label for each of the previous utterances ($u_1, ..., u_{i}$), i.e., the model has to predict a vector of labels of length $i$. Since EFR is a new task and has no direct baseline model, we extend the above ERC baselines to predict a vector of labels. We augment the output layer with $i$ independent softmax functions corresponding to each contextual utterance to achieve this. We keep the rest of the architecture as the original.

\subsection{\bf Experimental Setup}
We fine-tune BERT \cite{devlin2018bert} for ERC and extract its last layer hidden representation as utterance representation. We keep the standard vector dimension of BERT to represent an utterance. All reported results are averaged over 5 runs.
\subsubsection{\bf Hyperparameter Tuning} While training our models, we paid specific attention to tuning our hyperparameters appropriately. After manual tuning, the hyperparameters that we found to give us the best results are mentioned in Table \ref{tab:hyper}. For the masked memory network, the most important hyperparameter to tune was the number of memory hops. We experimented with hops ranging from 1 to 5. We observed that the performance kept increasing up to 4 hops and decreased when we increased the hops to 5. This can be seen in Table \ref{tab:hops}. For the Transformer-based network, an important hyperparameter was the number of encoder layers. We set the number of encoder layers to 6 as they gave the best results. Less than six layers gave a slightly lesser performance, whereas as we increased the number of layers from 6, the results more or less remained the same. This is illustrated in Table \ref{tab:encoderlayers}. Since we use BERT embeddings to represent utterances, the input hidden size for all our models is 768. The hidden dimension remains 768 until the start of classification layers. We used 3 and 5 linear layers for the ERC and EFR tasks for the masked memory networks, respectively. For the Transformer-based model, we employed a single linear layer as our classification layer for both tasks. The size of the hidden representations for all the models can be seen in Table \ref{tab:hyper}.

\begin{table}[h!]
\centering
{%
\begin{tabular}{|l|c|c|c|c|c|}
\hline
& \multicolumn{5}{c|}{\bf Hops} \\ \cline{2-6}
& 1 & 2 & 3 & \textcolor{blue}{4} & 5 \\ \hline
{\tt ERC-MMN} & 51.9 & 52.3 & 53.6 & \textcolor{blue}{55.7} & 54.9 \\ \hline
\end{tabular}%
}
\caption{Performance obtained by varying the number of hops in {\tt ERC-MMN}.}
\label{tab:hops}
\end{table}
\begin{table}[h!]
\centering
{%
\begin{tabular}{|l|c|c|c|c|c|c|c|c|}
\hline
& \multicolumn{8}{c|}{\bf Encoder layers} \\ \cline{2-9}
\multicolumn{1}{|c|}{} & 1 & 2 & 3 & 4 & 5 & \textcolor{blue}{6} & 7 & 8 \\ \hline
{\tt ERF-TX} & 33.1 & 37.7 & 40.2 & 41.4 & 42.6 & \textcolor{blue}{44.8} & 44.6 & 44.5 \\ \hline
\end{tabular}%
}
\caption{Performance obtained by varying the number of encoder layers in {\tt EFR-TX}.}
\label{tab:encoderlayers}
\end{table}
\begin{table*}[h!]
\centering
\resizebox{\textwidth}{!}{%
\begin{tabular}{l|c|c|c|c|c|c}
\multirow{2}{*}{\bf Hyperparameters} & \multicolumn{6}{c}{\bf Models} \\ \cline{2-7}
& {\tt ERC-MMN} & {\tt EFR-MMN} & {\tt ERC-TX} & {\tt EFR-TX} & \multitask
& \cascadeErcTrueEfr \\ \hline \hline
\textit{\textbf{Batch-size}} & 8 & 128 & 8 & 128 & 8
& 128 \\ \hline
\textit{\textbf{Learning rate}} & 1e-06 & 1e-07 & 5e-08 & 5e-08 & 1e-04
& 5e-07 \\ \hline
\textit{\textbf{Epochs}} & 100 & 100 & 1000 & 1000 & 100
& 1000 \\ \hline
\textit{\textbf{Dropout}} & 0.5 & 0.5 & 0.2 & 0.2 & 0.5
& 0.2 \\ \hline
\end{tabular}%
}
\caption{Hyperparameters used for our models. We set the number of hops in masked memory network ({\tt ERC-MMN}, {\tt EFR-MMN}, \multitask) as 4. The number of encoder layers in transformer-based networks ({\tt ERC-TX}, {\tt EFR-TX}, \cascadeErcTrueEfr) is set at 6. The input vector size for all models is 768.}
\label{tab:hyper}
\end{table*}
\begin{table}[h!]
\centering
{%
\begin{tabular}{l|c|c}
\multirow{2}{*}{\bf System} & \multicolumn{2}{c}{\bf Classification FC layers} \\ \cline{2-3}
& \bf \#Layer & \bf Dimensions \\ \hline \hline
{\tt ERC-MMN} & \multirow{2}{*}{\em 3} & \multirow{2}{*}{\em 1536$\rightarrow$768$\rightarrow$384$\rightarrow$7} \\ 
\multitask {(\tt ERC)} &  & \\ \hline

{\tt EFR-MMN} & \multirow{2}{*}{\em 5} & \multirow{2}{*}{\em 1536$\rightarrow$1536$\rightarrow$1536$\rightarrow$768$\rightarrow$384$\rightarrow$2} \\ 
\multitask {(\tt EFR)} &  &  \\ \hline

{\tt ERC-TX} & \textit{1} & \textit{1536$\rightarrow$7} \\ \hline

{\tt EFR-TX} & \multirow{2}{*}{\em 1} & \multirow{2}{*}{\em 1536$\rightarrow$2} \\ 
\cascadeErcTrueEfr & & \\ \hline
\end{tabular}%
}
\caption{Number of classification FC layers used in our models along with the hidden vector size.}
\label{tab:fclayers}
\end{table}
Another important hyperparameter in conversational analysis systems is the sequence length of a dialogue, i.e., the maximum number of utterances in a dialogue. For the ERC task, we consider the sequence length as 15 as only a couple of dialogues contain more than 15 utterances. For the EFR task, we set this sequence length equal to 5. We carefully selected this sequence length as more than 95\% triggers lie in this distance range from the target utterance. This phenomenon can be seen in Figure \ref{fig:eda_trig} of the main text.

Other hyperparameters such as batch size, learning rate, dropout, and maximum epochs were also tuned manually after observing the model's performance with a range of values for them. For instance, if the learning rate for {\tt ERC-MMN} was set more than $1e^{-6}$, the learning was in a zig-zag fashion, whereas a lower learning rate slowed the learning to a large extent. A similar trend was observed towards the learning rate for the other models too. If we look at the maximum number of epochs mentioned in Table \ref{tab:hyper}, we can see that the masked memory network converged faster than the Transformer-based model. This may be due to the increased parameters in the Transformer model. We mention the maximum epochs required for our models to converge. Some may reach the best performance before these many epochs with the mentioned learning rate. We use dropouts to reduce the effect of overfitting.

We show the number of FC layers used in our models for classification in Table \ref{tab:fclayers}. We also mention the size for the input and output vectors. We set these values after careful considerations and several experiments.

\subsection{\bf Experimental Results}
In this section, we present our experimental results and comparative study for both ERC and EFR tasks. \subsubsection{\bf Single-task Learning Framework} In this setup, both tasks are trained and evaluated separately. We evaluate two tasks on both Transformer-based ({\tt TX}) and masked memory network-based ({\tt MMN}) architectures. Table~\ref{tab:erc:results} summarizes the results. The MMN based systems, i.e., {\tt ERC-MMN} and {\tt EFR-MMN} (jointly denoted as {\tt (ERC/EFR)-MMN} in Table \ref{tab:erc:results}), obtain F1-scores of 55.78\% and 33.42\%, respectively. The modeling of {\tt EFR-MMN} as utterance-level classification follows the same procedure adopted for the baselines of EFR (c.f. Baseline section). Though the {\tt MMN} architecture yields moderate performance on ERC, it underperforms on the EFR task, possibly due to the way the task was modeled. This motivated us to model EFR as an instance-level classification, as mentioned in Table~\ref{tab:instance:EFR} and the methodology section. Subsequently, we train a Transformer-based architecture {\tt EFR-TX} and obtain a 44.79\% F1-score on the test set. The improvement of more than 11\% in F1-score justifies our EFR modeling as instance-level classification. We argue that reasoning the flip requires the information of the emotional states of the speakers. To support this hypothesis, we propose the \cascadeErcTrueEfr\  architecture, where we provide the true emotion labels in the {\tt EFR-TX} architecture. The results obtained from this model support our hypothesis as we obtain a 53.9\% F1-score. That is, an improvement of 9\% over the {\tt EFR-TX} model is observed. Following the success of {\tt EFR-TX}, we also experiment with {\tt ERC-TX}; however, the performance degrades significantly.

\begin{table*}[t]
\centering
\renewcommand*{\arraystretch}{1.1}
\resizebox{\textwidth}{!}
{%
\begin{tabular}{l||c|c|c|c|c|c|c|c||c}
\multirow{2}{*}{\bf System} & \multicolumn{8}{c||}{\textbf{ERC (F1)}} & \textbf{EFR (F1)} \\ \cline{2-10}
& \textbf{Dg} & \textbf{Jy} & \textbf{Sr} & \textbf{An} & \textbf{Fr} & \textbf{Ne} & \textbf{Sa} & \textbf{W-Avg} & \textbf{Trigger} \\ \hline \hline

CMN$^\dagger$ & 0.0 & 48.6 & \textbf{54.0} & 33.7 & 8.6 & \textbf{75.9} & 19.9 & 51.7 & 37.5 \\ 

ICON$^\dagger$ & 0.0 & 36.8 & 45.5 & 37.0 & 0.0 & 69.6 & 11.0 & 50.1 & 37.3 \\

DGCN$^\dagger$ & 0.0 & 48.1 & 52.9 & 31.6 & 4.5 & 75.8 & 15.5 & 51.8 & 52.9 \\
DGCN$_{\text{multi}}^\dagger$ & 0.0 & 39.2 & 43.7 & 37.1 & 0.0 & 71.7 & 12.1 & 51.1 & 53.0 \\

AGHMN$^\dagger$ & 0.0 & 40.1 & 43.1 & 11.7 & 0.0 & 63.0 & 25.0 & 44.2 & 52.3 \\ 
Pointer Network$^\dagger$ & 3.0 & 15.1 & 17.0 & 13.1 & 0.0 & 63.2 & 7.0 & 35.1 & 49.0 \\ \hline 

{\tt (ERC/EFR)-MMN} & \textbf{20.2} & \textbf{48.7} & 50.4 & 42.9 & 9.8 & 71.9 & 29.6 & \textbf{55.7} & 33.4 \\ 
{\tt (ERC/EFR)-TX} & 0.0 & 4.0 & 5.0 & 1.9 & 0.0 & 61.2 & 0.0 & 29.5 & 44.8 \\ \hline

\multitask & 18.8 & 48.6 & 49.3 & \textbf{43.7} & \textbf{11.2} & 72.1 & \textbf{32.0} & \textbf{55.7} & 34.8 \\ \hline

\cascadeErcTrueEfr & - & - & - & - & - & - & - & - & \textbf{53.9} \\ \hline 

\end{tabular}%
}
\caption{Comparative analysis for ERC and EFR. {\tt (ERC/EFR)-MMN} represents {\tt ERC-MMN} for the emotion recognition task and {\tt EFR-MMN} for the emotion-flip reasoning task (Dg: disgust, Jy: joy, Sr: surprise, An: anger, Fr: fear, Ne: neutral, Sa: sadness). $^\dagger$ \textbf{Performance on the MELD-FR dataset}.
}
\label{tab:erc:results}
\end{table*}

\subsubsection{\bf Multitask Learning Framework}
Since ERC and EFR depend on each other, we design a joint learning approach where two tasks are learned simultaneously in a unified manner.
{This setting follows a pipeline where the {\tt ERC-MMN} model is extended to first detect the ERC labels, and if there is a flip observed in a speaker's emotion, the EFR task is performed. Both the tasks share {\tt ERC-MMN} till the penultimate layer.}
Subsequently, we add two parallel fully-connected layers -- one for each task. We train the model by accumulating the losses incurred at the output layers of ERC and EFR and backpropagate
employing {\tt Adam} optimizer. We call this model \multitask.

Unfortunately, \multitask\ does not benefit much for ERC and a slight improvement of $\sim1.4\%$ F1-score for EFR compared to {\tt EFR-MMN}. However, the obtained multitask performance on  EFR  is below-par compared to {\tt EFR-TX}.
Similar to the earlier case, we attribute this performance drop to the different ways in which we model the tasks.
We also perform multitasking on the best baseline, DGCN, where we observe that EFR slightly improves (from $52.9\%$ to $53.0\%$), whereas we observe a $0.7\%$ drop in ERC ($51.8\%$ to $51.1\%$).

\subsubsection{\bf Other Cascade Models}
Along with the experiments mentioned in Table \ref{tab:erc:results}, we also perform other experiments to play around with our architectures, specifically for the EFR task. In Section \ref{sec:exp}, we explain two types of cascade models for the EFR task ({\cascadeErcEfr} and {\cascadeErcTrueEfr}). Here, we will show two more cascade models that we tried for the EFR task.

\begin{enumerate}
    \item \textbf{Early-fusion cascade:} In this setting, we introduce emotion labels in the input layer of our model. We concatenate the emotion representation (a 7-dimensional one-hot vector) with the utterance representation (a 768-dimensional BERT vector) and then feed it to the transformer-based network. The first row of Table \ref{tab:extraExpResults} shows the results obtained using this model.
    
    \item \textbf{Late-fusion cascade:} In this setting, we introduce emotion labels in the penultimate layer of our model. We concatenate the emotion representation (a 7-dimensional one-hot vector) with the representation obtained from the transformer {encoder}. We then feed this representation to the classification layers for classification. The second row of Table \ref{tab:extraExpResults} shows the results obtained using this model.
\end{enumerate}


\begin{table}[h!]
\centering
{%
\begin{tabular}{l|l|l}
 & \textbf{\em Early Fusion} & \textbf{\em Late Fusion} \\ \hline
\textbf{Trigger F1} & \multicolumn{1}{c|}{35.1} & \multicolumn{1}{c}{51.5} \\ \hline
\end{tabular}%
}
\caption{Experimental results for early- and late-fusion of emotion labels in the \cascadeErcTrueEfr\ model.}
\label{tab:extraExpResults}
\end{table}

\subsection{\bf Comparative Analysis}
We utilize the publicly available implementations of the baselines for the comparative study and report the performance in Table \ref{tab:erc:results}. For ERC, DGCN turns out to be the best baseline ($51.80\%$), while CMN ($51.70\%$) and ICON ($50.15\%$) yield comparable performances. Pointer network seems to be the worst performing baseline ($35.1\%$). In comparison, our proposed single-task system, {\tt ERC-MMN} reports the best performance with an improvement of $\sim4\%$ against DGCN with $55.78\%$ F1-score.

In the EFR task, we obtain mixed results. In comparison to the single-task model {\tt ERC-TX} with $44.79\%$ F1, two of the baselines (CMN with $37.5\%$ F1 and ICON with $37.3\%$ F1) obtain lesser F1-scores, while the other three baselines, DGCN, AGHMN, and pointer network report improved results of  $52.93\%$, $52.30\%$, and $49.0\%$ F1-scores, respectively. However, all five baselines are outperformed by our final model (\cascadeErcTrueEfr ) which reports an improvement of $1-17\%$ in F1-score for the trigger label.

Another critical observation is that due to class-imbalance, all baselines find it difficult to identify the \textit{disgust} emotion. Similarly, three out of five baselines fail to classify any \textit{fear} emotion as well. In contrast, our proposed models correctly identify both \textit{disgust} and \textit{fear} emotions for at least a few utterances. Furthermore, except for \textit{surprise} and \textit{neutral}, our proposed model outperforms the baselines in remaining five emotion labels.      

{As shown in Table \ref{tab:erc:results}, the higest performance for ERC is achieved by our model which contains memory network along with stacked GRUs and masked attention. We also obtain similar results for the multitask framework for ERC. We believe this performance boost is the result of the use of memory network in an effective manner. As can be seen from the baseline results, AGHMN and Pointer Network do not perform at par with the others since they do not contain any memory component. While CMN, ICON and our method perform better. Moreover, we notice that the ERC task performs in a comparative fashion in both standalone and multitask settings. On the other hand, our model performs better for the EFR task when emotions are being learnt simultaneously. This suggests that the ERC task assists the EFR task.}

{
\subsection{\bf Generalizability}
To analyze the performance of our model on an out-of-distribution generalization test set, we consider another dataset, IEMOCAP  \cite{busso2008iemocap}. It contains crowdsourced conversations revolving around $16$ topics. For the construction of our test set, we randomly pick two conversations from each topic. We then create instances from these dialogues as illustrated in Table \ref{tab:instance:EFR} and manually annotate them with EFR labels (inter-annotator agreement, $\alpha = 0.813$)\footnote{{The same annotation process as described in Section \ref{sec:dataset} was followed to annotate IEMOCAP. {$\alpha_{AB}=0.818$, $\alpha_{AC}=0.808$, and $\alpha_{BC}=0.820$. To measure the overall agreement score, we take the average of these values, $\alpha=0.813$.}
}}. Table \ref{tab:iemocap_stats} gives us a brief statistics of the IEMOCAP-FR dataset. We test our model trained on MELD-FR on IEMOCAP-FR and report the results in Table \ref{tab:erc:results:iemocap}.
For ERC, our model produces the best results. However, the results are significantly less than the results obtained on MELD-FR. This reduction can be attributed to the inherent differences in the dialogues present in the two datasets. IEMOCAP contains more than 50 utterances in a dialogue on average whereas MELD contains an average of 9 utterances per dialogue. Secondly, the emotion distribution between the two sets also differ in a major way. IEMOCAP does not contain any \textit{disgust} emotion, and the \textit{neutral} emotion is not as commonly present in it as it is in MELD-FR. Consequently, the task of emotion recognition becomes challenging for this dataset. On the other hand, our model and the baselines perform surprisingly well for the task of EFR, comparable to the EFR performance on MELD-FR. This performance can be attributed to the fact that even if the emotion distribution differs in IEMOCAP-FR, the distribution of triggers is still very similar to MELD-FR.

\begin{table}[t]
\centering
\resizebox{\textwidth}{!}{%
{\begin{tabular}{|c|c|c|c|c|c|c|c|c|c|c|}
\hline
\multicolumn{8}{|c|}{\textbf{ERC}} & \multicolumn{3}{c|}{\textbf{EFR}} \\ \hline
\textbf{Disgust} & \multicolumn{1}{c|}{\textbf{Joy}} & \multicolumn{1}{c|}{\textbf{Surprise}} & \multicolumn{1}{c|}{\textbf{Anger}} & \multicolumn{1}{c|}{\textbf{Fear}} & \multicolumn{1}{c|}{\textbf{Neutral}} & \multicolumn{1}{c|}{\textbf{Sadness}} & \multicolumn{1}{c|}{\textbf{Total}} & \textbf{\#Diag with Flip} & \multicolumn{1}{c|}{\textbf{\#Utt with Flip}} & \multicolumn{1}{c|}{\textbf{\#Triggers}} \\ \hline \hline
0 & 671 & 44 & 1413 & 25 & 552 & 407 & 3112 & 32 & 965 & 1388 \\ \hline
\end{tabular}}%
}
\caption{Statistics of the IEMOCAP-FR dataset.}
\label{tab:iemocap_stats}
\end{table}
\begin{table*}[t]
\centering
\renewcommand*{\arraystretch}{1.1}
\resizebox{\textwidth}{!}
{%
\begin{tabular}{l||c|c|c|c|c|c|c|c||c}
\multirow{2}{*}{\bf System} & \multicolumn{8}{c||}{\textbf{ERC (F1)}} & \textbf{EFR (F1)} \\ \cline{2-10}
& \textbf{Dg} & \textbf{Jy} & \textbf{Sr} & \textbf{An} & \textbf{Fr} & \textbf{Ne} & \textbf{Sa} & \textbf{W-Avg} & \textbf{Trigger} \\ \hline \hline

CMN & 0.0 & 7.1 & 0.0 & \textbf{56.4} & 0.0 & 2.1 & 2.3 & 28.2 & 35.6\\ 

ICON & 0.0 & 14.2 & 0.0 & 49.7 & 0.0 & 9.1 & 8.0 & 28.4 & 36.1 \\

DGCN & 0.0 & 15.5 & 2.3 & 54.2 & 0.0 & 6.0 & 11.5 & 30.8 & 49.6 \\
DGCN$_{\text{multi}}$ & 0.0 & 11.3 & 2.2 & 53.4 & 0.0 & 5.1 & 13.1 & 29.2 & 48.4\\

AGHMN & 0.0 & 7.3 & 0.0 & 45.8 & 0.0 & 0.0 & 11.2 & 24.4 & 49.3 \\ 
Pointer Network & 0.0 & 12.4 & 0.0 & 32.2 & 0.0 & 2.6 & 6.0 & 18.1 & 44.7 \\ \hline 

{\tt (ERC/EFR)-MMN} & 0.0 & \textbf{19.3} & \textbf{3.2} & 52.7 & 0.0 & \textbf{10.2} & \textbf{17.1} & \textbf{33.7} & 31.2 \\ 
{\tt (ERC/EFR)-TX} & 0.0 & 11.9 & 1.3 & 36.5 & 0.0 & 4.2 & 9.5 & 21.2 & 40.1 \\ \hline

\multitask & 0.0 & 17.5 & 2.2 & 51.5 & 0.0 & 8.3 & \textbf{17.7} & \textbf{31.4} & 32.8 \\ \hline

\cascadeErcTrueEfr & - & - & - & - & - & - & - & - & \textbf{52.8} \\ \hline 

\end{tabular}%
}
\caption{Comparative analysis for ERC and EFR. {\tt (ERC/EFR)-MMN} represents {\tt ERC-MMN} for the emotion recognition task and {\tt EFR-MMN} for the emotion-flip reasoning task (Dg: disgust, Jy: joy, Sr: surprise, An: anger, Fr: fear, Ne: neutral, Sa: sadness). \textbf{Trained on MELD-FR; Tested on IEMOCAP-FR dataset}.}

\label{tab:erc:results:iemocap}
\end{table*}
}

\subsection{\bf Error Analysis}
This section presents both quantitative and qualitative analysis \textit{w.r.t.} the confusion matrix and misclassification examples. We also supplement our analysis of the proposed systems with DGCN (the best baseline). Tables \ref{tab:erc:confusion} and \ref{tab:efr:confusion} show the confusion matrices for ERC and EFR, respectively. 
In the ERC task, for most of the emotions, our proposed {\tt EFR-MMN} model reports better true-positives against the baseline, as highlighted in blue-colored text in Table \ref{tab:erc:confusion}. 

\begin{table}[t!]
\centering
\resizebox{\columnwidth}{!}{%
\begin{tabular}{l|l|c:c:c:c:c:c:c|}
\multicolumn{1}{c}{} & \multicolumn{1}{c}{} & \multicolumn{7}{c}{\bf Predicted} \\ \cline{3-9}
\multicolumn{1}{c}{} & & \multicolumn{1}{c|}{\textbf{Disgust}} & \multicolumn{1}{c|}{\textbf{Joy}} & \multicolumn{1}{c|}{\textbf{Surprise}} & \multicolumn{1}{c|}{\textbf{Anger}} & \multicolumn{1}{c|}{\textbf{Fear}} & \multicolumn{1}{c|}{\textbf{Neutral}} & \multicolumn{1}{c|}{\textbf{Sadness}} \\ \cline{2-9} 

\multirow{7}{*}{\rotatebox{90}{\bf Actual}} & \textbf{Disgust}  & \cellcolor{red!10} \textcolor{blue}{15/0} & 5/5     & 4/10   & 14/9   & 1/0  & 19/33   & 3/4 \\ \cline{2-2} \cdashline{3-9}
& \textbf{Joy}      & 13/0 & \cellcolor{red!10} \textcolor{blue}{157/110} & 12/32  & 26/19  & 4/0  & 104/154 & 9/10 \\ \cline{2-2} \cdashline{3-9}
& \textbf{Surprise} & 7/0  & 31/40   & \cellcolor{red!10} \textcolor{blue}{115/87} & 32/21  & 4/0  & 44/80   & 5/10 \\ \cline{2-2} \cdashline{3-9} 
& \textbf{Anger}    & 16/0 & 32/45   & 31/56  & \cellcolor{red!10} \textcolor{blue}{118/74} & 6/0  & 64/101  & 16/7 \\ \cline{2-2} \cdashline{3-9}
& \textbf{Fear}     & 1/0  & 4/5     & 5/9    & 7/2    & \cellcolor{red!10} \textcolor{blue}{4/0}  & 14/24   & 7/3 \\ \cline{2-2} \cdashline{3-9} 
& \textbf{Neutral}  & 28/0 & 72/54   & 42/44  & 50/19  & 14/0 & \cellcolor{red!10} \textcolor{red}{705/808} & 32/18 \\ \cline{2-2} \cdashline{3-9}
& \textbf{Sadness}  & 7/0  & 18/25   & 9/12  & 19/17  & 6/0  & 68/100  & \cellcolor{red!10} \textcolor{blue}{42/15} \\ \cline{2-9}
\end{tabular}%
}
\caption{Confusion matrices for ERC. Cell (a/b) represents `$a$' number of samples predicted by {\tt EFR-MMN} (our best model) and `$b$' number of samples predicted by DGCN. \textcolor{blue}{Blue-colored} and \textcolor{red}{red-colored} texts represent \textcolor{blue}{superiority} and \textcolor{red}{inferiority} of  {\tt EFR-MMN}, respectively, compared to the baseline, {\tt DGCN}.}
\label{tab:erc:confusion}
\end{table}

The confusion matrix of {\tt EFR-MMN} reveals the most confusing pair of emotions as \textit{neutral} and \textit{joy}, with $104$ \textit{joy} samples misclassified as \textit{neutral} and $72$ \textit{neutral} samples misclassified as \textit{joy}.  
Another important observation is that DGCN ignores the under-represented emotions, such as \textit{disgust} and \textit{fear}, completely. In contrast, our proposed model assigns these two emotions to a few utterances with a bit of success. It suggests  {\tt EFR-MMN} to be unbiased towards the under-represented emotion labels. 

In EFR, the baseline DGCN reports comparable performance with \cascadeErcTrueEfr. However, while analysing the confusion matrices in Table \ref{tab:efr:confusion}, we observe that the number of \textit{true-positives}  considering the \textit{trigger} class is much higher for \cascadeErcTrueEfr\ (926) than DGCN (627). Also, \cascadeErcTrueEfr\ reports lesser \textit{false-negatives}. On the other hand, however, the \textit{false-positives} are more, and due to which our proposed model fails to leverage the higher \textit{true-positives} to a full extent and reports only $1\%$ improvement over DGCN.

\begin{table}[t!]
\centering

\subfloat[\label{tab:efr:confusion:proposed}{\cascadeErcTrueEfr}]
{
\resizebox{0.45\columnwidth}{!}{%
\begin{tabular}{l|l|c:c|}

\multicolumn{1}{c}{} & \multicolumn{1}{c}{} & \multicolumn{2}{c}{\bf Predicted} \\ \cline{3-4}

\multicolumn{1}{c}{} & \multicolumn{1}{c|}{} & \multicolumn{1}{c|}{\textbf{Non-Trigger}} & \multicolumn{1}{c|}{\textbf{Trigger}} \\ \cline{2-4} 
 
\multirow{2}{*}{\rotatebox{90}{\bf Actual}} & \textbf{Non-Trigger} & \cellcolor{red!10} 2144 & 1359 \\ \cline{2-2} \cdashline{3-4}

& \textbf{Trigger} & 226 & \cellcolor{red!10} \textcolor{blue}{926}  \\ \cline{2-4} 
\end{tabular}%
}}
\hspace{0.5em}
\subfloat[\label{tab:efr:confusion:drnn}Baseline: DGCN]
{
\resizebox{0.45\columnwidth}{!}{%
\begin{tabular}{l|l|c:c|}

\multicolumn{1}{c}{} & \multicolumn{1}{c}{} & \multicolumn{2}{c}{\bf Predicted} \\ \cline{3-4}

\multicolumn{1}{c}{} & \multicolumn{1}{c|}{} & \multicolumn{1}{c|}{\textbf{Non-Trigger}} & \multicolumn{1}{c|}{\textbf{Trigger}} \\ \cline{2-4} 
 
\multirow{2}{*}{\rotatebox{90}{\bf Actual}} & \textbf{Non-Trigger} & \cellcolor{red!10} 2913 & 590 \\ \cline{2-2} \cdashline{3-4}

& \textbf{Trigger} & 525 & 627  \\ \cline{2-4} 
\end{tabular}%
}}
\caption{Confusion matrix of our best model and the best baseline for the EFR task.}
\label{tab:efr:confusion}
\end{table}
\begin{table*}[ht!]
\centering
\resizebox{\textwidth}{!}
{%
\begin{tabular}{l|l|p{25em}|c|c|c}
\multicolumn{1}{c|}{} & \multicolumn{1}{c|}{} & \multicolumn{1}{c|}{} & & \multicolumn{2}{c}{\textbf{Prediction}} \\ \cline{5-6} 
\multicolumn{1}{c|}{\multirow{-2}{*}{\textbf{\#}}} & \multicolumn{1}{c|}{\multirow{-2}{*}{\textbf{Speaker}}} & \multicolumn{1}{c|}{\multirow{-2}{*}{\textbf{Utterance}}} & \multirow{-2}{*}{\textbf{Actual}} & {\tt ERC-MMN} & DGCN \\ \hline \hline

$u_1$ & Phoebe & Well alright! We already tried feeding her, changing her, burping her. Oh! Try this one. & \textit{sadness} & {\color[HTML]{FE0000}\textit{anger}} &  {\color[HTML]{FE0000}\textit{joy}} \\ \hline

$u_2$ & Phoebe & Go back in time and listen to Phoebe! & \textit{anger} & \textit{anger} & {\color[HTML]{FE0000}\textit{joy}} \\ \hline

\multirow{2}{*}{$u_3$} & \multirow{2}{*}{Monica} & Alright here's something. It says to try holding the baby close to your body and then swing her rapidly from side to side. & \multirow{2}{*}{\textit{neutral}} & \multirow{2}{*}{\textit{neutral}} & \multirow{2}{*}{\textit{neutral}} \\ \hline

$u_4$ & Rachel & Ok. & \textit{neutral} & \textit{neutral} & \textit{neutral} \\ \hline

$u_5$ & Monica & It worked! & \textit{surprise} & \textit{surprise} & {\color[HTML]{FE0000} \textit{anger}} \\ \hline

$u_6$ & Rachel & Oh! Oh! No, just stopped to throw up a little bit. & \textit{sadness} & \textit{sadness} & {\color[HTML]{FE0000}\textit{neutral}} \\ \hline

$u_7$ & Rachel & Oh come on! What am I gonna do? Its been hours and it won't stop crying! & \textit{sadness} & \textit{sadness} & {\color[HTML]{FE0000}\textit{neutral}} \\ \hline

$u_8$ & Monica & Umm 'she' Rach not 'it', 'she'. & \textit{neutral} & \textit{neutral} & \textit{neutral} \\ \hline

$u_9$ & Rachel & Yeah I'm not so sure. & \textit{neutral} & \textit{neutral} & \textit{neutral} \\ \hline

$u_{10}$ & Monica & Oh my god! I am losing my mind! & \textit{anger} & \textit{anger} & \textit{anger} \\ \hline

\end{tabular}%
}
\caption{Actual and predicted emotions for a dialogue having 10 utterances ($u_1,...,u_{10}$) from the test set. \textcolor{red}{Red-colored} text represents misclassification. For the given example, {\tt ERC-MMN} misclassifies only one utterance, whereas DGCN (best baseline) commits mistakes for 5 out of 10 utterances.}
\label{tab:erc:error}
\end{table*}
\begin{table*}[ht!]
\centering
\resizebox{\textwidth}{!}
{%
\begin{tabular}{l|l|p{13em}|l|c|c|c|c}
\multicolumn{1}{c|}{} & & \multicolumn{2}{c|}{} & \multicolumn{2}{c|}{\bf Actual} & \multicolumn{2}{c}{\textbf{Prediction}} \\ \cline{5-8} 
\multicolumn{1}{c|}{} & & \multicolumn{2}{c|}{}  & \textbf{Emotion} & \textbf{Trigger} & & \\ 
\multicolumn{1}{c|}{\multirow{-2}{*}{\textbf{\#}}} & \multicolumn{1}{c|}{\multirow{-2}{*}{\textbf{Speaker}}} & \multicolumn{2}{c|}{\multirow{-2}{*}{\textbf{Utterance}}} & \textbf{(MELD)} & \textbf{(MELD-FR)} & \multicolumn{1}{c|}{\multirow{-2}{*}{\cascadeErcTrueEfr}} & \multicolumn{1}{c}{\multirow{-2}{*}{DGCN}} \\ \hline \hline

$u_1$ & Ross & Okay & \multirow{4}{*}{Context}& \textit{neutral} & \textit{N-Trigger} & \textit{N-Trigger} & \textit{N-Trigger} \\ \cline{1-3} \cline{5-8}

$u_2$ & Rachel & Ross didn’t you say that there was an elevator in here? & & \textcolor{green!40!black!90}{\bf \textit{neutral}} & \textit{N-Trigger} & \textit{N-Trigger} & \textit{N-Trigger} \\ \cline{1-3} \cline{5-8}

$u_3$ & Ross & Uhh yes I did but there isn’t okay here we go! & & \textit{sadness} & \textit{Trigger} & \textit{Trigger} & {\color[HTML]{FE0000} \textit{N-Trigger}} \\ \cline{1-3} \cline{5-8}

$u_4$ & Ross & Okay go left left left & & \textit{surprise} & \textit{Trigger} & \textit{Trigger} & \textit{Trigger} \\ \hline \hline

$u_5$ & Rachel & Okay y’know what there is no more left left! & Target & \textcolor{blue}{\bf \textit{anger}} & \textit{N-Trigger} & {\color[HTML]{FE0000} \textit{Trigger}} & {\color[HTML]{FE0000} \textit{Trigger}} \\ \hline

\end{tabular}%
}
\begin{tikzpicture}[remember picture, overlay]
      \node[draw, ellipse, minimum height=0.1em, minimum width=4.0em, blue, line width=0.4mm] (anger) [xshift=2.1em, yshift=1.9em] {};
      \node[draw, ellipse, minimum height=0.1em, minimum width=4.0em, green!40!black!90, line width = 0.4mm] (neutral) [xshift=2.1em, yshift=8.25em] {};
      \path[line width=0.3mm, ->, magenta] (neutral.west) edge [bend right=88] (anger.west);
      \node[magenta] (flip) [xshift=-0.6em, yshift=6.0em] {Flip};
      
    \end{tikzpicture}
\caption{Actual and predicted labels of triggers for a dialogue having five utterances ($u_1,...,u_{5}$) from the test set. There is an emotion-flip for Rachel (\textit{\textcolor{green!40!black!90}{neutral}}$\rightarrow$\textit{\textcolor{blue}{anger}}) in $u_5$ and its triggers are $u_3$ and $u_4$. We mark them as triggers because Ross tricked her into believing that his apartment had an elevator and still acted like nothing happened, thus instigating an emotion-flip. 
}
\label{tab:efr:error}
 \end{table*}

We also perform error analysis on the predictions of proposed systems. For illustration, we present one representative dialogue with its gold and predicted labels (ours and DGCN) for each task.
We observe from Table \ref{tab:erc:error} that in a dialogue of ten utterances with three speakers, {\tt EFR-MMN} misclassifies only one utterance, whereas DGCN misclassifies five utterances in the same dialogue. 
Moreover, DGCN predicts six utterances
as \textit{neutral}, out of which only four are correct. In comparison, our proposed model does not misclassify any emotion as \textit{neutral} in the dialogue. It can be related to the biasness of DGCN towards the majority emotions.

Similarly, Table \ref{tab:efr:error} shows a dialogue for EFR. There are two speakers, Ross and Rachel, and we observe an emotion-flip (\textit{neutral}$\rightarrow$\textit{anger}) for Rachel in utterance $u_5$ considering her previous utterance $u_2$. For the target utterance $u_5$, actual trigger utterances are $u_3$ and $u_4$. Our proposed model, \cascadeErcTrueEfr, correctly identifies both triggers; however, it also misidentifies one utterance, i.e., $u_5$, as trigger. For the same dialogue, DGCN misclassifies one \textit{trigger} utterance as \textit{non-trigger} and one \textit{non-trigger} utterance as \textit{trigger}. 
\section{Conclusion}\label{sec:conclusion}
A change in speaker's emotion is usually conditioned on either internal (within conversation) or external (outside conversation) factors. This paper focused on the internal factors and defined Emotion-Flip Reasoning (EFR) in conversations that highlight all the contextual instigators responsible for an emotion-flip -- a first study of its kind in the conversational dialogue.

Since emotion recognition (ERC) and EFR are closely related, we addressed both of them in the current study. We proposed a masked memory network and a Transformer-based architecture for the ERC and EFR tasks, respectively. Further, we leveraged the interaction between the two tasks through a multitask framework.
As a by-product of the study, we developed a new benchmark dataset, MELD-FR, for the EFR task. It is an extension of MELD emotion recognition dataset with trigger labels for each emotion flips. In total, we manually annotated MELD-FR with 8,500 trigger utterances for 5,400 emotion-flips. We performed an extensive evaluation on the MELD-FR dataset and compared the experimental results with various existing state-of-the-art systems. Results suggest that the proposed system obtained better performance as compared to the state-of-the-art.

\noindent \paragraph{\bf Use Cases and Future Directions}
One of the objectives of ERC is to develop an empathetic dialogue system, where the knowledge of emotion can assist the dialogue agent in generating empathetic responses; thus, the user enjoys a more natural conversation. Our proposed EFR task is the next step towards the same objective. The information of emotion-flips can be viewed as a feedback mechanism to the dialogue agent. If the dialogue agent senses a flip of emotion during the conversation and has the information of possible reason (trigger) of that flip, it can refine its subsequent response. For example, if it encounters a negative emotion-flip (neutral to anger), it can refrain from generating a similar (trigger) response to avoid negative emotion-flip. On the other hand, for a positive emotion-flip (anger to joy), the confidence of the dialogue system in generating such responses will increase in the future.        

EFR opens a new thread in the explainability of the speaker's emotion-dynamics in conversation, and our effort is just a baby step towards this direction. In the future, we would like to explore emotion-flip reasoning in more detail. Instead of considering the contextual utterances as possible triggers, we can define a label set to capture the extent of appraisals. We will explore this direction in the future. {Additionally, we will explore the effect of multimodal signals like audio and video in detecting emotion flips.}

\section*{Acknowledgement}
T. Chakraborty would like to acknowledge the support of Ramanujan Fellowship, CAI, IIIT-Delhi and ihub-Anubhuti-iiitd Foundation set up under the NM-ICPS scheme of the Department of Science and Technology, India.

\bibliography{references}

\begin{thebibliography}{10}
\expandafter\ifx\csname url\endcsname\relax
  \def\url#1{\texttt{#1}}\fi
\expandafter\ifx\csname urlprefix\endcsname\relax\def\urlprefix{URL }\fi
\expandafter\ifx\csname href\endcsname\relax
  \def\href#1#2{#2} \def\path#1{#1}\fi

\bibitem{Ekman92anargument}
P.~Ekman, {An argument for basic emotions}, {Cognition and Emotion} (1992)
  169--200.

\bibitem{Picard:1997:AC:265013}
R.~W. Picard, {Affective Computing}, MIT Press, Cambridge, MA, USA, 1997.

\bibitem{abdul-mageed-ungar-2017-emonet}
M.~Abdul-Mageed, L.~Ungar, {E}mo{N}et: Fine-grained emotion detection with
  gated recurrent neural networks, in: Proceedings of the 55th Annual Meeting
  of the Association for Computational Linguistics (Volume 1: Long Papers),
  Vancouver, Canada, 2017, pp. 718--728.

\bibitem{chatterjee-etal-2019-semeval}
A.~Chatterjee, K.~N. Narahari, M.~Joshi, P.~Agrawal, {S}em{E}val-2019 task 3:
  {E}mo{C}ontext contextual emotion detection in text, in: Proceedings of the
  13th International Workshop on Semantic Evaluation, Minneapolis, Minnesota,
  USA, 2019, pp. 39--48.

\bibitem{akhtar:all:in:one:affective}
M.~S. {Akhtar}, D.~{Ghosal}, A.~{Ekbal}, P.~{Bhattacharyya}, S.~{Kurohashi},
  All-in-one: Emotion, sentiment and intensity prediction using a multi-task
  ensemble framework, IEEE Transactions on Affective Computing (2019) 1--1.

\bibitem{gupta-etal-2010-emotion}
N.~Gupta, M.~Gilbert, G.~Di~Fabbrizio, Emotion detection in email customer
  care, in: Proceedings of the {NAACL} {HLT} 2010 Workshop on Computational
  Approaches to Analysis and Generation of Emotion in Text, Los Angeles, CA,
  2010, pp. 10--16.

\bibitem{dini-bittar-2016-emotion}
L.~Dini, A.~Bittar, Emotion analysis on twitter: The hidden challenge, in:
  Proceedings of the Tenth International Conference on Language Resources and
  Evaluation ({LREC}'16), Portoro{\v{z}}, Slovenia, 2016, pp. 3953--3958.

\bibitem{akhtar:cim:emotion}
M.~S. {Akhtar}, A.~{Ekbal}, E.~{Cambria}, How intense are you? predicting
  intensities of emotions and sentiments using stacked ensemble [application
  notes], IEEE Computational Intelligence Magazine 15~(1) (2020) 64--75.

\bibitem{khanpour-caragea-2018-fine}
H.~Khanpour, C.~Caragea, Fine-grained emotion detection in health-related
  online posts, in: Proceedings of the 2018 Conference on Empirical Methods in
  Natural Language Processing, Brussels, Belgium, 2018, pp. 1160--1166.

\bibitem{hazarika-etal-2018-conversational}
D.~Hazarika, S.~Poria, A.~Zadeh, E.~Cambria, L.-P. Morency, R.~Zimmermann,
  Conversational memory network for emotion recognition in dyadic dialogue
  videos, in: Proceedings of the 2018 Conference of the North {A}merican
  Chapter of the Association for Computational Linguistics: Human Language
  Technologies, Volume 1 (Long Papers), New Orleans, Louisiana, 2018, pp.
  2122--2132.

\bibitem{lin2019moel}
Z.~Lin, A.~Madotto, J.~Shin, P.~Xu, P.~Fung, Moel: Mixture of empathetic
  listeners, arXiv preprint arXiv:1908.07687 (2019).

\bibitem{shin2020generating}
J.~Shin, P.~Xu, A.~Madotto, P.~Fung, Generating empathetic responses by looking
  ahead the user’s sentiment, in: ICASSP 2020-2020 IEEE International
  Conference on Acoustics, Speech and Signal Processing (ICASSP), IEEE, 2020,
  pp. 7989--7993.

\bibitem{ma2020survey}
Y.~Ma, K.~L. Nguyen, F.~Z. Xing, E.~Cambria, A survey on empathetic dialogue
  systems, Information Fusion 64 (2020) 50--70.

\bibitem{young2020dialogue}
T.~Young, V.~Pandelea, S.~Poria, E.~Cambria, Dialogue systems with audio
  context, Neurocomputing 388 (2020) 102--109.

\bibitem{end2end:mem:net:subakhtar:nips:2015}
S.~Sukhbaatar, A.~Sszlam, J.~Weston, R.~Fergus, End-to-end memory networks, in:
  Proceedings of the Advances in Neural Information Processing Systems, 2015,
  pp. 2440--2448.

\bibitem{poria-etal-2019-meld}
S.~Poria, D.~Hazarika, N.~Majumder, G.~Naik, E.~Cambria, R.~Mihalcea, {MELD}: A
  multimodal multi-party dataset for emotion recognition in conversations, in:
  Proceedings of the 57th Annual Meeting of the Association for Computational
  Linguistics, Florence, Italy, 2019, pp. 527--536.

\bibitem{lee-etal-2010-text}
S.~Y.~M. Lee, Y.~Chen, C.-R. Huang, A text-driven rule-based system for emotion
  cause detection, in: Proceedings of the {NAACL} {HLT} 2010 Workshop on
  Computational Approaches to Analysis and Generation of Emotion in Text, Los
  Angeles, CA, 2010, pp. 45--53.

\bibitem{gui-etal-2016-event}
L.~Gui, D.~Wu, R.~Xu, Q.~Lu, Y.~Zhou, Event-driven emotion cause extraction
  with corpus construction, in: Proceedings of the 2016 Conference on Empirical
  Methods in Natural Language Processing, Austin, Texas, 2016, pp. 1639--1649.

\bibitem{poria2016deeper}
S.~Poria, E.~Cambria, D.~Hazarika, P.~Vij, A deeper look into sarcastic tweets
  using deep convolutional neural networks, arXiv preprint arXiv:1610.08815
  (2016).

\bibitem{MENCATTINI201468}
A.~Mencattini, E.~Martinelli, G.~Costantini, M.~Todisco, B.~Basile, M.~Bozzali,
  C.~{Di Natale},
  \href{https://www.sciencedirect.com/science/article/pii/S0950705114001087}{Speech
  emotion recognition using amplitude modulation parameters and a combined
  feature selection procedure}, Knowledge-Based Systems 63 (2014) 68--81.
\newblock \href {https://doi.org/https://doi.org/10.1016/j.knosys.2014.03.019}
  {\path{doi:https://doi.org/10.1016/j.knosys.2014.03.019}}.
\newline\urlprefix\url{https://www.sciencedirect.com/science/article/pii/S0950705114001087}

\bibitem{ZHANG2016248}
L.~Zhang, K.~Mistry, S.~C. Neoh, C.~P. Lim,
  \href{https://www.sciencedirect.com/science/article/pii/S0950705116302799}{Intelligent
  facial emotion recognition using moth-firefly optimization}, Knowledge-Based
  Systems 111 (2016) 248--267.
\newblock \href {https://doi.org/https://doi.org/10.1016/j.knosys.2016.08.018}
  {\path{doi:https://doi.org/10.1016/j.knosys.2016.08.018}}.
\newline\urlprefix\url{https://www.sciencedirect.com/science/article/pii/S0950705116302799}

\bibitem{CUI2020106243}
H.~Cui, A.~Liu, X.~Zhang, X.~Chen, K.~Wang, X.~Chen,
  \href{https://www.sciencedirect.com/science/article/pii/S0950705120304433}{Eeg-based
  emotion recognition using an end-to-end regional-asymmetric convolutional
  neural network}, Knowledge-Based Systems 205 (2020) 106243.
\newblock \href {https://doi.org/https://doi.org/10.1016/j.knosys.2020.106243}
  {\path{doi:https://doi.org/10.1016/j.knosys.2020.106243}}.
\newline\urlprefix\url{https://www.sciencedirect.com/science/article/pii/S0950705120304433}

\bibitem{hazarika-etal-2018-icon}
D.~Hazarika, S.~Poria, R.~Mihalcea, E.~Cambria, R.~Zimmermann, {ICON}:
  Interactive conversational memory network for multimodal emotion detection,
  in: Proceedings of the 2018 Conference on Empirical Methods in Natural
  Language Processing, Brussels, Belgium, 2018, pp. 2594--2604.

\bibitem{hazarika2019emotion}
D.~Hazarika, S.~Poria, R.~Zimmermann, R.~Mihalcea, Emotion recognition in
  conversations with transfer learning from generative conversation modeling,
  arXiv preprint arXiv:1910.04980 (2019).

\bibitem{zhong-etal-2019-knowledge}
P.~Zhong, D.~Wang, C.~Miao, Knowledge-enriched transformer for emotion
  detection in textual conversations, in: Proceedings of the 2019 Conference on
  Empirical Methods in Natural Language Processing and the 9th International
  Joint Conference on Natural Language Processing (EMNLP-IJCNLP), Hong Kong,
  China, 2019, pp. 165--176.

\bibitem{ghosal-etal-2019-dialoguegcn}
D.~Ghosal, N.~Majumder, S.~Poria, N.~Chhaya, A.~Gelbukh, {D}ialogue{GCN}: A
  graph convolutional neural network for emotion recognition in conversation,
  in: Proceedings of the 2019 Conference on Empirical Methods in Natural
  Language Processing and the 9th International Joint Conference on Natural
  Language Processing (EMNLP-IJCNLP), Hong Kong, China, 2019, pp. 154--164.

\bibitem{li2020bieru}
W.~Li, W.~Shao, S.~Ji, E.~Cambria, Bieru: bidirectional emotional recurrent
  unit for conversational sentiment analysis, arXiv preprint arXiv:2006.00492
  (2020).

\bibitem{zhang2019quantum}
Y.~Zhang, Q.~Li, D.~Song, P.~Zhang, P.~Wang, Quantum-inspired interactive
  networks for conversational sentiment analysis. (2019).

\bibitem{wang2020contextualized}
Y.~Wang, J.~Zhang, J.~Ma, S.~Wang, J.~Xiao, Contextualized emotion recognition
  in conversation as sequence tagging, in: Proceedings of the 21th Annual
  Meeting of the Special Interest Group on Discourse and Dialogue, 2020, pp.
  186--195.

\bibitem{lazarus1984stress}
R.~S. Lazarus, S.~Folkman, Stress, appraisal, and coping, 1984.

\bibitem{krippendorff2011computing}
K.~Krippendorff, Computing krippendorff's alpha-reliability (2011).

\bibitem{vaswani2017attention}
A.~Vaswani, N.~Shazeer, N.~Parmar, J.~Uszkoreit, L.~Jones, A.~N. Gomez,
  {\L}.~Kaiser, I.~Polosukhin, Attention is all you need, in: Advances in
  neural information processing systems, 2017, pp. 5998--6008.

\bibitem{IAN:2017}
D.~Ma, S.~Li, X.~Zhang, H.~Wang, Interactive attention networks for
  aspect-level sentiment classification, in: Proceedings of the Twenty-Sixth
  International Joint Conference on Artificial Intelligence, {IJCAI} 2017,
  Melbourne, Australia, 2017, pp. 4068--4074.

\bibitem{jiao2019real}
W.~Jiao, M.~R. Lyu, I.~King, Real-time emotion recognition via attention gated
  hierarchical memory network, arXiv preprint arXiv:1911.09075 (2019).

\bibitem{vinyals2015pointer}
O.~Vinyals, M.~Fortunato, N.~Jaitly, Pointer networks, arXiv preprint
  arXiv:1506.03134 (2015).

\bibitem{devlin2018bert}
J.~Devlin, M.-W. Chang, K.~Lee, K.~Toutanova, Bert: Pre-training of deep
  bidirectional transformers for language understanding, arXiv preprint
  arXiv:1810.04805 (2018).

\bibitem{busso2008iemocap}
C.~Busso, M.~Bulut, C.-C. Lee, A.~Kazemzadeh, E.~Mower, S.~Kim, J.~N. Chang,
  S.~Lee, S.~S. Narayanan, Iemocap: Interactive emotional dyadic motion capture
  database, Language resources and evaluation 42~(4) (2008) 335--359.

\end{thebibliography}

\end{document}